\documentclass{article}
\usepackage{arxiv}

\usepackage{natbib}

\usepackage[utf8]{inputenc} %
\usepackage[T1]{fontenc}    %
\usepackage{hyperref}       %
\hypersetup{
  colorlinks=true,
  citecolor=MidnightBlue,
}
\usepackage[normalem]{ulem} 
\usepackage{url}            %
\usepackage{booktabs}       %
\usepackage{amsfonts}       %
\usepackage{nicefrac}       %
\usepackage{pifont}
\usepackage{microtype}      %
\usepackage{xcolor}         %
\usepackage{amsmath}
\usepackage[utf8]{inputenc} %
\usepackage[T1]{fontenc}    %
\usepackage{url}            %
\usepackage{adjustbox}
\usepackage{caption}
\usepackage{booktabs}       %
\usepackage{amsfonts}       %
\usepackage{nicefrac}       %
\usepackage{microtype}      %
\usepackage[capitalize]{cleveref}
\usepackage{tabularx}
\usepackage{longtable}

\usepackage{amsmath}
\usepackage{pifont}
\usepackage{graphicx}
\usepackage[dvipsnames,table]{xcolor}
\usepackage{bbding}
\newcommand{\redcross}{\textcolor{red}{\XSolidBrush}}
\newcommand{\greencheck}{\textcolor{Green}{\CheckmarkBold}}

\usepackage{xspace}
\usepackage{makecell}  %
\usepackage{enumitem}
\usepackage{multirow}
\usepackage{float}
\usepackage{arydshln}
\usepackage{multirow}
\usepackage{caption}
\usepackage{subcaption}
\usepackage{algpseudocode}
\usepackage{siunitx}
\usepackage{wrapfig}

\usepackage{rotating}

\usepackage[ruled,vlined,linesnumbered,noend]{algorithm2e}
\SetKwInput{KwIn}{Input}
\SetKwInput{KwOut}{Output}
\SetKwComment{tcp}{// }{}
\DontPrintSemicolon

\usepackage{soul}
\usepackage{threeparttable}

\begin{document}

\title{When Does Multimodal Learning Help in Healthcare?\\
A Benchmark on EHR and Chest X-Ray Fusion}

\renewcommand{\shorttitle}{A Benchmark on EHR and Chest X-Ray Fusion}

\author{%
    Kejing Yin\thanks{Corresponding author.}\\
    Department of Computer Science\\
    Hong Kong Baptist University\\
    \texttt{cskjyin@comp.hkbu.edu.hk}
    \And
    Haizhou Xu\\
    Department of Computer Science\\
    Hong Kong Baptist University\\
    \texttt{cshzxu@comp.hkbu.edu.hk}
    \And
    Wenfang Yao\\
    Division of Artificial Intelligence\\
    Lingnan University\\
    \texttt{dorothyyao@ln.edu.hk}
    \And
    Chen Liu\\
    School of Nursing\\
    The Hong Kong Polytechnic University\\
    \texttt{laura-chen.liu@connect.polyu.hk}
    \And
    Zijie Chen\\
    Department of Computer Science\\
    Hong Kong Baptist University\\
    \texttt{cszjchen@comp.hkbu.edu.hk}
    \And
    Yui Haang Cheung\\
    Department of Computer Science\\
    Hong Kong Baptist University\\
    \texttt{21223270@life.hkbu.edu.hk}
    \And
    William K. Cheung\\
    Department of Computer Science\\
    Hong Kong Baptist University\\
    \texttt{william@comp.hkbu.edu.hk}
    \And
    Jing Qin\\
    School of Nursing\\
    The Hong Kong Polytechnic University\\
    \texttt{harry.qin@polyu.edu.hk}
    }

\date{}

\maketitle

\begin{abstract}
Machine learning holds promise for advancing clinical decision support, yet it remains unclear when multimodal learning truly helps in practice, particularly under modality missingness and fairness constraints. In this work, we conduct a systematic benchmark of multimodal fusion between Electronic Health Records (EHR) and chest X-rays (CXR) on standardized cohorts from MIMIC-IV and MIMIC-CXR, aiming to answer four fundamental questions: when multimodal fusion improves clinical prediction, how different fusion strategies compare, how robust existing methods are to missing modalities, and whether multimodal models achieve algorithmic fairness. 
Our study reveals several key insights. Multimodal fusion improves performance when modalities are complete, with gains concentrating in diseases that require complementary information from both EHR and CXR. While cross-modal learning mechanisms capture clinically meaningful dependencies beyond simple concatenation, the rich temporal structure of EHR introduces strong modality imbalance that architectural complexity alone cannot overcome. Under realistic missingness, multimodal benefits rapidly degrade unless models are explicitly designed to handle incomplete inputs. Moreover, multimodal fusion does not inherently improve fairness, with subgroup disparities mainly arising from unequal sensitivity across demographic groups. 
To support reproducible and extensible evaluation, we further release a flexible benchmarking toolkit that enables plug-and-play integration of new models and datasets. Together, this work provides actionable guidance on when multimodal learning helps, when it fails, and why, laying the foundation for developing clinically deployable multimodal systems that are both effective and reliable. The open-source toolkit can be found at https://github.com/jakeykj/CareBench.
\end{abstract}

\section{Introduction}
Machine learning is increasingly transforming clinical decision-making, with models capable of forecasting disease onset~\citep{venugopalan2021multimodal,el2020multimodal}, stratifying patient risk~\citep{boehm2022multimodal}, and personalizing treatment pathways~\citep{esteva2022prostate}. A key frontier in this domain is multimodal learning~\citep{elsharief25medmod}, which aims to create a holistic patient view by integrating heterogeneous data sources such as Electronic Health Records (EHR) and Chest X-rays (CXR). The fusion of rich longitudinal EHR data with critical diagnostic CXR imaging has shown great potential for improved accuracy in many downstream prediction tasks~\citep{mlhc2022hayatmedfuse,yao2024drfuse}.

Recent benchmarking efforts have advanced reproducible evaluation in clinical machine learning, including YAIB~\citep{vandewater24yet} for unimodal EHR modeling and MedMod~\citep{elsharief25medmod} for paired multimodal data. However, existing benchmarks primarily emphasize aggregate predictive performance under complete-modality assumptions, offering limited insight into when multimodal learning truly helps in practice. In particular, the field lacks systematic understanding of how fusion strategies behave under modality imbalance and missing data, as well as how multimodal models affect subgroup disparities. While fairness has been studied in medical imaging~\citep{zong23medfair}, these questions remain largely unexplored in the context of multimodal EHR–CXR fusion.

To address these limitations, we present \textbf{CareBench}, a comprehensive benchmark for multimodal fusion of EHR and chest X-rays. Our work provides three core components to the research community: (i) an open-source and reproducible \textbf{data extraction pipeline} for MIMIC-IV and MIMIC-CXR to establish a standard cohort for evaluation; (ii) a unified, open-source, and reproducible \textbf{toolkit} implementing a wide array of models, from unimodal baselines to state-of-the-art fusion architectures, to facilitate fair comparison and future extensions; and (iii) a rigorous \textbf{evaluation protocol} that moves beyond standard predictive performance. Critically, this framework enables systematic analyses of model behavior under varying degrees of modality missingness and across different patient subgroups, supporting a principled investigation of robustness and algorithmic fairness in multimodal EHR–CXR fusion.

Through this benchmark, we show that multimodal fusion improves performance when modalities are complete, with gains concentrating in diseases that require complementary information from both EHR and imaging. We further demonstrate that while cross-modal learning captures clinically meaningful dependencies beyond simple concatenation, EHR’s rich temporal structure introduces strong modality imbalance that architectural complexity alone cannot overcome. Under realistic missingness, multimodal benefits rapidly diminish unless models are explicitly designed to handle incomplete inputs. Finally, we find that multimodal fusion does not inherently improve algorithmic fairness, with subgroup disparities primarily arising from unequal sensitivity across demographic groups. Together, these findings clarify when multimodal learning helps, when it fails, and why, providing actionable guidance for developing clinically deployable multimodal systems.

\section{Related Work}
Prior benchmarks have enabled systematic comparison of clinical learning methods, yet most focus on either unimodal settings or a restricted range of multimodal fusion approaches. As shown in Table~\ref{tab:comparison}, existing efforts rarely support extensible evaluation across unimodal and multimodal models under diverse data conditions. Consequently, fundamental questions about when multimodal learning helps, when it fails, and how fusion strategies interact with missing data and subgroup disparities remain insufficiently explored.

\paragraph{EHR Benchmarks in Healthcare}
The widespread adoption of electronic health records (EHRs) has enabled the creation of large-scale datasets, spurring the development of numerous benchmarks for clinical prediction tasks. 
\citet{purushothamBenchmarkingDeepLearning2018} and \citet{harutyunyanMultitaskLearningBenchmarking2019} introduced early EHR benchmarks on MIMIC-III, demonstrating the utility of deep learning models in clinical outcome prediction. 
\citet{barbieriBenchmarkingDeepLearning2020} extended this line by evaluating neural ODEs and attention-based models for readmission and patient risk stratification. 
\citet{sheikhalishahiBenchmarkingMachineLearning2020} compared machine learning models on the multi-center eICU dataset, highlighting generalization across healthcare systems. 
To address data accessibility, MIMIC-Extract~\citep{wangMIMICExtractDataExtraction2020} and FIDDLE~\citep{tangDemocratizingEHRAnalyses2020} provided standardized preprocessing pipelines, while Clairvoyance~\citep{jarrettCLAIRVOYANCEPIPELINETOOLKIT2021} offered an end-to-end AutoML-friendly framework for medical time-series. 
More recent efforts include EHR-TS-PT~\citep{mcdermott2021comprehensive} and EHRSHOT~\citep{wornow2023ehrshot}, which explored pre-training and few-shot learning for EHR time series, as well as HiRID-ICU~\citep{yecheHiRIDICUBenchmarkComprehensiveMachine2022}, which benchmarked machine learning models on high-resolution ICU data. 
Finally, YAIB~\citep{vandewater24yet} proposed a modular, multi-dataset EHR framework emphasizing extensibility. 
Despite their contributions, all these efforts focus exclusively on the EHR modality, whereas real-world clinical decision-making is inherently multimodal.

\paragraph{Multimodal Benchmarks for Clinical Prediction}
Recognizing the benefits of integrating multimodal data for clinical tasks, recent years have seen a growing number of multimodal benchmarks. 
INSPECT~\citep{huang2023inspect} and RadFusion~\citep{zhou2021radfusion} established multimodal benchmarks for pulmonary embolism diagnosis and prognosis using CT and EHR data, though both adopted only late-fusion strategies. 
MC-BEC~\citep{chen2023multimodal} introduced a multimodal benchmark for emergency care with EHR, notes, and waveforms, and uniquely assessed robustness to missing data and fairness. However, its fusion approach was limited to a simple late-fusion scheme, leaving advanced and adaptive fusion strategies unexplored. 
PyHealth~\citep{yangPyHealthDeepLearning2023} provided a comprehensive deep learning toolkit covering EHR, waveforms, text, and imaging; however, it does not provide standardized performance comparisons across models. 
Most relevant to our work, MedMod~\citep{elsharief25medmod} introduced the first EHR–CXR benchmark, comparing early, joint, and late fusion paradigms. Yet, MedMod did not systematically evaluate robustness to missing modalities or fairness across subgroups. 
In parallel, MEDFAIR~\citep{zong23medfair} focused on fairness benchmarking in imaging, but was limited to unimodal settings. 
Together, these works underscore the need for a more comprehensive benchmark.

\paragraph{Multimodal Fusion for Clinical Prediction}
A growing body of work has explored multimodal fusion methods for clinical prediction, addressing challenges such as heterogeneous data distributions, irregular sampling, and modality imbalance. Simple late fusion remains a common baseline, while more advanced approaches—including DAFT~\citep{daft-polsterl2021combining}, MMTM~\citep{joze2020mmtm}, and UTDE~\citep{zhang2023improving}, enable tighter cross-modal interactions under complete-modality settings. 
To address modality absence, several methods introduce explicit mechanisms for flexible fusion, such as mixture-of-experts designs, disentangled representations, or latent-space imputation, including HEALNet~\citep{hemker2024healnet}, Flex-MoE~\citep{yun2024flex}, DrFuse~\citep{yao2024drfuse}, UMSE~\citep{lee2023learning}, and M3Care~\citep{zhang2022m3care}. While these approaches differ substantially in their modeling assumptions and treatment of missing modalities, they are often evaluated under limited or inconsistent data conditions.
As a result, it remains unclear when sophisticated fusion mechanisms truly outperform simpler baselines, how they interact with modality imbalance and missing data, and whether their benefits generalize across clinical tasks and patient subgroups, motivating a systematic benchmark of multimodal fusion methods.

\section{Dataset Extraction}

We constructed our benchmark using large-scale real-world ICU databases, specifically MIMIC-IV~\citep{johnson2023mimic} and MIMIC-CXR~\citep{johnson2019mimic}. The former contains de-identified records of adult patients admitted to either intensive care units or the emergency department of Beth Israel Deaconess Medical Center (BIDMC) between 2008 and 2019, and the latter is a publicly available dataset of chest radiographs collected from BIDMC, where a subset of patients can be matched with those in MIMIC-IV. 

\subsection{Cohort Construction}

We construct two cohorts of ICU stays from the MIMIC-IV database: a base cohort containing all ICU episodes that satisfy clinical and temporal consistency requirements, and a matched subset further restricted to encounters with paired chest radiographs. The detailed exclusion criteria for constructing the data cohorts can be found in \cref{fig:exclusion_criteria} in the Appendix.

\paragraph{The Base Cohort} Starting from the 73,181 ICU stays available from the MIMIC-IV database, we remove stays lacking essential clinical documentation (e.g., missing discharge notes or diagnostic codes) and episodes with implausible temporal records, such as hospital admission times occurring after ICU admission or discharge. To focus on clinically meaningful acute episodes, we further excluded ICU stays of less than 6 hours, and admissions labeled as non-urgent or elective, repeated ICU episodes within the same hospitalization. Since short ICU stays often represent observational or step-down care, and usually have insufficient longitudinal information for robust prediction, we further exclude ICU stays shorter than 48 hours to construct the base cohort, which eventually contains 26,947 ICU stays.

\paragraph{The Matched Subset}To establish a multimodal benchmark, we require the availability of at least one chest radiograph within a window spanning 24 hours before to 48 hours after ICU admission. This yields a matched subset of 7,149 ICU stays, representing patients for whom both structured EHR data and chest radiographs are available. 

\subsection{Feature Extraction}
\paragraph{EHR Feature Extraction}

We extracted a comprehensive set of structured electronic health record (EHR) features from the MIMIC-IV v2.2 database. Our extraction pipeline was designed to capture clinically relevant variables across multiple physiological domains, including vital signs, neurological status (Glasgow Coma Scale), cardiac rhythm, respiratory support parameters (O$_2$ flow, FiO$_2$), fluid balance (urine output), and body weight. All features were retrieved using structured SQL queries executed on a locally deployed PostgreSQL instance of MIMIC-IV v2.2, with the Python toolkit SQLAlchemy for database interaction.  To ensure temporal alignment within each ICU stay, we joined relevant source tables, including chartevents, labevents, procedureevents, as well as derived modules such as gcs,  kdigo\_uo,  ventilator\_setting,  blood\_differential,  weight\_durations, and  enzyme, using  stay\_id and timestamp synchronization. We initially explored 25 distinct categories of measurements. However, features with a missingness rate greater than 90\% were empirically excluded. Furthermore, treatment-related variables, such as continuous renal replacement therapy (CRRT), invasive line placement, and mechanical ventilation settings, were removed to avoid potential label leakage. The final set of features used in CareBench is summarized in Table~\ref{tab:features}.

\paragraph{EHR Preprocessing}
To maintain compatibility with most baseline models, and following prior MIMIC benchmarks~\citep{harutyunyanMultitaskLearningBenchmarking2019, elsharief25medmod}, we resampled the EHR data at an hourly resolution. Missing values were imputed using forward filling and median imputation strategies, while binary cardiac rhythm indicators were directly imputed with 0. To preserve information on data availability, we additionally retained binary mask columns indicating the presence or absence of each measurement, as missingness itself can be informative in clinical settings~\citep{morid2023time}. For continuous variables, we applied robust normalization using the median and interquartile range (IQR) to mitigate the influence of outliers.

\paragraph{CXR Selection Criteria}
To ensure temporal and clinical alignment between imaging and EHR data, we restricted the chest X-ray (CXR) cohort to scans acquired during the patient’s current ICU stay. Only frontal-view images with an Anterior-Posterior (AP) projection were included, as this is the standard acquisition protocol for bedside radiography in critical care settings. Among all eligible AP views, we selected the most recent  CXR prior to the prediction timepoint to best reflect the patient’s latest cardiopulmonary status.

\begin{figure}
    \includegraphics[width=\linewidth]{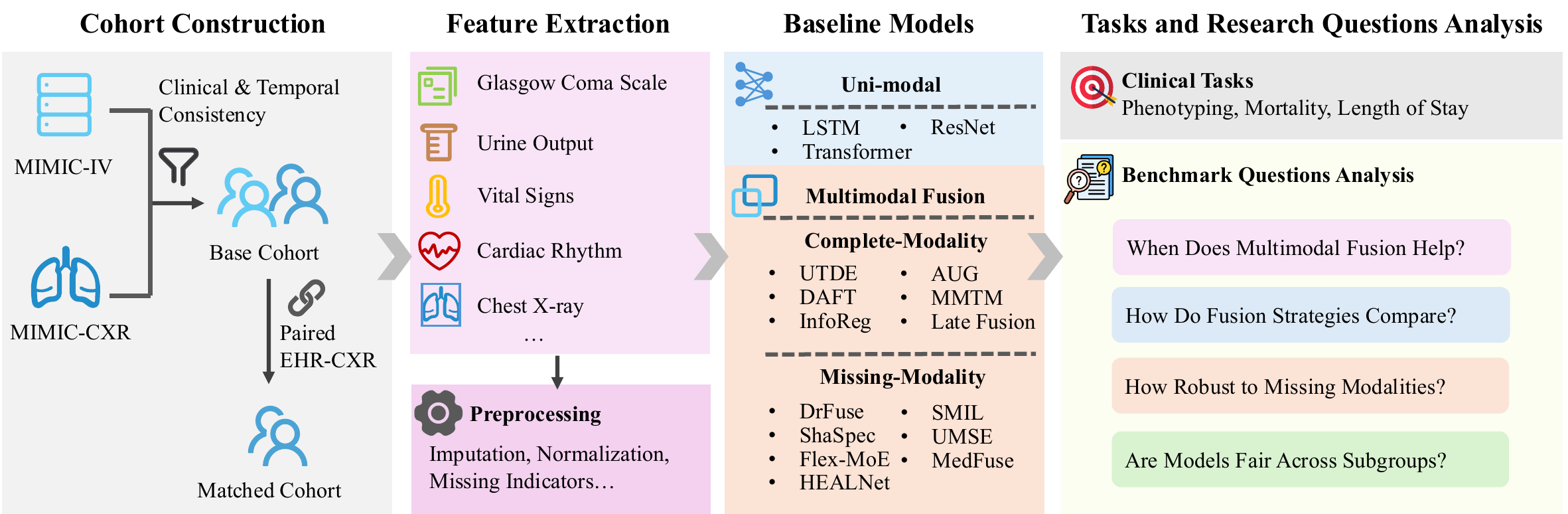}
    \caption{Overview of CareBench pipeline.}
    \label{fig:framework}
\end{figure}

\section{Benchmark Design}
\subsection{Models}

We benchmark a broad set of models for multimodal fusion of EHR and chest X-rays, spanning unimodal baselines, simple fusion strategies, and recent state-of-the-art multimodal algorithms that can be adapted to clinical settings. Detailed description of these baselines can be found in \cref{app:models}.

\paragraph{Uni-modal Baselines} We include uni-modal models as baselines to establish reference performance for each modality. For EHR, we include the classic Long Short-Term Memory network (LSTM) and the Transformer model. They are widely used for capturing temporal dependencies in sequential EHR data. For CXR, we adopt the ResNet-50 model, which is pretrained on ImageNet. These baselines quantify the stand-alone predictive value of each modality.
\paragraph{Complete-Modality Multimodal Fusion Methods}  This group of models assumes that all modalities are present at both training and inference,
including Late Fusion, UTDE~\citep{zhang2023improving}, DAFT~\citep{daft-polsterl2021combining}, MMTM~\citep{joze2020mmtm}, AUG~\citep{Jiang2025aug} and InfoReg~\citep{huang2025adaptive}.
\paragraph{Missing-Modality Multimodal Fusion Methods} We implement a broad set of multimodal fusion methods that could handle missing modalities, covering both models developed specifically for clinical data and models originally proposed in other domains (e.g., video–audio classification) that can be naturally adapted to clinical EHR–CXR fusion. This collection spans diverse design paradigms, including: HEALNet~\citep{hemker2024healnet}, Flex-MoE~\citep{yun2024flex}, DrFuse~\citep{yao2024drfuse}, UMSE~\citep{lee2023learning}, ShaSpec~\citep{wang2023multi}, M3Care~\citep{zhang2022m3care}, MedFuse~\citep{mlhc2022hayatmedfuse}, and SMIL~\citep{ma2021smil}.

\subsection{Downstream Tasks and Evaluations}

We evaluate models on three downstream tasks that are highly relevant to clinical decision support: phenotyping classification, mortality prediction, and length-of-stay (LoS) prediction. All tasks use patient data observed within a fixed prediction window, and are evaluated on both the matched subset (complete modalities) and the base cohort (realistic setting with missing modalities). To ensure comparability, we adopt patient-level train/validation/test splits and report established metrics tailored to each task.

\paragraph{Phenotyping Classification}
The goal of phenotyping is to predict the set of acute and chronic conditions present during an ICU stay. Following prior benchmarks, we construct 25 phenotypes derived from ICD-9 and ICD-10 diagnosis codes, spanning common comorbidities and critical conditions. The task is formulated as multi-label classification, requiring models to output a binary prediction for each phenotype simultaneously. To comprehensively evaluate model performance, we employ a suite of metrics including Area Under the Receiver Operating Characteristic Curve (AUROC), Area Under the Precision-Recall Curve (AUPRC), F1 score, precision, recall, specificity, and accuracy (ACC).

\paragraph{Mortality Prediction}
This task focuses on predicting in-hospital mortality within the first 48 hours of ICU admission, formulated as a binary classification problem. The objective is to determine whether a patient will survive or die during hospitalization, enabling early identification of critically ill patients at high risk of deterioration. Labels are derived directly from hospital discharge status, with positive cases defined as patients who died during their hospital stay and negative cases as those who were discharged alive. To ensure clinically realistic evaluation, the prediction window is restricted to the initial 48 hours of an ICU stay, using only information available within that period. Model performance is assessed using a comprehensive set of metrics, including AUROC and AUPRC to capture threshold-independent discrimination ability, as well as F1-score, accuracy (ACC), precision, recall, and specificity.

\paragraph{LoS Prediction}
Accurate estimation of ICU LoS is important for clinical planning and resource allocation. In this task, we use the first 48 hours of EHR and CXR data to predict the remaining hospital stay (RLOS). The RLOS is discretized into clinically meaningful intervals: 2–3 days, 3–4 days, 4–5 days, 5–6 days, 6–7 days, 7–14 days, and 14+ days, resulting in a multi-class classification problem with ordinal structure. Performance is evaluated using the ACC, F1 score, and Cohen's Kappa weighted quadratic, with additional metrics such as precision, recall, and specificity reported for completeness.

\subsection{Implementation Details}

We implement all models in Python 3.12.2 using PyTorch 2.5.1 and PyTorch Lightning 2.2, running with CUDA 12.1 and cuDNN 9.1.0. Experiments are conducted on servers equipped with AMD EPYC 7763 64-Core CPUs, 512 GB RAM, and 4×NVIDIA RTX 4090 GPUs (24 GB memory each). To ensure fair comparison, we adopt consistent training settings across tasks and perform Bayesian hyperparameter optimization to tune model-specific configurations.

\paragraph{Hyperparameter tuning}
We employ Bayesian optimization with a Gaussian process surrogate (gp-minimize from scikit-optimize) and the GP-Hedge acquisition strategy, which adaptively balances exploitation and exploration by combining multiple acquisition functions (LCB, EI, PI). Each search runs for 20 iterations, starting from 5 random initial configurations. For each candidate configuration, the framework launches full training runs 
with three random seeds 
to account for variance. Results from all seeds are aggregated by computing the mean and standard deviation of multiple metrics (ACC, AUPRC, AUROC, F1, etc.), and the task-specific selection criterion is applied: we maximize AUPRC for phenotyping and mortality, and ACC for LoS prediction. 

\paragraph{Hyperparameter search space}
To ensure fair comparison, we standardize the core training setup across all models (learning rate, batch size, epochs, early stopping, etc.) and restrict hyperparameter search to model-specific components. Thus, hyperparameter search is only applied to models with explicit model-specific parameters, namely DrFuse, Flex-MoE, HEALNet, M3Care, ShaSpec, SMIL, AUG and InfoReg. An overview of their searched parameters and ranges is provided in Table~\ref{tab:search_space_overview}.

\section{Benchmark Results and Discussions}

\begin{table}\footnotesize
    \centering
    \caption{Predictive performance over the matched subset. \sethlcolor{pink!50}\hl{Pink-highlighted cells} indicate the best uni-modal method for each disease.
\sethlcolor{green!20}\hl{Green-highlighted cells} denote results that are statistically significantly better than the best uni-modal method based on a permutation test ($p<0.05$).
Boldface values indicate the best-performing model for each evaluation metric (column).}
    \label{tab:prediction_matched}
    \resizebox{\textwidth}{!}{
    \setlength{\tabcolsep}{3pt}
    \begin{tabular}{lcccccccccc}
    \toprule
     & \multicolumn{3}{c}{\textbf{Phenotyping}} & \multicolumn{3}{c}{\textbf{Mortality}} & \multicolumn{3}{c}{\textbf{Length of Stay}} \\\cmidrule(lr){2-4}\cmidrule(lr){5-7}\cmidrule(lr){8-10}
     Model & AUPRC & AUROC & F1 & AUPRC & AUROC & F1 & Accuracy & F1 & Kappa \\
    \midrule
    \textbf{Transformer} & \cellcolor{pink!50}0.472{\footnotesize$\pm$0.0008} & \cellcolor{pink!50}0.724{\footnotesize$\pm$0.0005} & \cellcolor{pink!50}0.340{\footnotesize$\pm$0.0123} & 0.447{\footnotesize$\pm$0.0151} & \cellcolor{pink!50}0.821{\footnotesize$\pm$0.0036} & 0.576{\footnotesize$\pm$0.0341} & 0.392{\footnotesize$\pm$0.0101} & \cellcolor{pink!50}0.192{\footnotesize$\pm$0.0070} & 0.187{\footnotesize$\pm$0.0106}\\
    \textbf{LSTM} & 0.452{\footnotesize$\pm$0.0036} & 0.715{\footnotesize$\pm$0.0057} & 0.328{\footnotesize$\pm$0.0227} & \cellcolor{pink!50}0.451{\footnotesize$\pm$0.0243} & 0.803{\footnotesize$\pm$0.0261} & \cellcolor{pink!50}0.627{\footnotesize$\pm$0.0597} & \cellcolor{pink!50}0.397{\footnotesize$\pm$0.0044} & 0.186{\footnotesize$\pm$0.0136} & \cellcolor{pink!50}0.192{\footnotesize$\pm$0.0060}\\
    \textbf{ResNet} & 0.400{\footnotesize$\pm$0.0014} & 0.668{\footnotesize$\pm$0.0022} & 0.269{\footnotesize$\pm$0.0145} & 0.225{\footnotesize$\pm$0.0062} & 0.688{\footnotesize$\pm$0.0090} & 0.492{\footnotesize$\pm$0.0224} & 0.346{\footnotesize$\pm$0.0076} & 0.133{\footnotesize$\pm$0.0025} & 0.115{\footnotesize$\pm$0.0022}\\\midrule
    \textbf{LateFusion} & \cellcolor{green!20}0.489{\footnotesize$\pm$0.0007} & \cellcolor{green!20}0.738{\footnotesize$\pm$0.0012} & 0.298{\footnotesize$\pm$0.0182} & 0.433{\footnotesize$\pm$0.0179} & 0.823{\footnotesize$\pm$0.0039} & 0.611{\footnotesize$\pm$0.0443} & 0.394{\footnotesize$\pm$0.0068} & 0.189{\footnotesize$\pm$0.0094} & 0.191{\footnotesize$\pm$0.0053}\\
    \textbf{MedFuse} & 0.478{\footnotesize$\pm$0.0030} & 0.732{\footnotesize$\pm$0.0014} & 0.256{\footnotesize$\pm$0.0191} & 0.472{\footnotesize$\pm$0.0098} & \bf 0.851{\footnotesize$\pm$0.0056} & 0.548{\footnotesize$\pm$0.0454} & 0.403{\footnotesize$\pm$0.0013} & 0.167{\footnotesize$\pm$0.0078} & 0.194{\footnotesize$\pm$0.0035}\\
    \textbf{MMTM} & 0.472{\footnotesize$\pm$0.0048} & 0.727{\footnotesize$\pm$0.0012} & \bf \cellcolor{green!20}0.381{\footnotesize$\pm$0.0116} & 0.357{\footnotesize$\pm$0.0540} & 0.801{\footnotesize$\pm$0.0323} & 0.621{\footnotesize$\pm$0.0571} & 0.261{\footnotesize$\pm$0.0062} & 0.165{\footnotesize$\pm$0.0069} & 0.091{\footnotesize$\pm$0.0089}\\
    \textbf{DAFT} & \cellcolor{green!20}0.480{\footnotesize$\pm$0.0034} & 0.731{\footnotesize$\pm$0.0034} & 0.275{\footnotesize$\pm$0.0217} & 0.435{\footnotesize$\pm$0.0131} & 0.834{\footnotesize$\pm$0.0050} & 0.472{\footnotesize$\pm$0.0056} & 0.394{\footnotesize$\pm$0.0030} & 0.160{\footnotesize$\pm$0.0133} & 0.180{\footnotesize$\pm$0.0087}\\
    \textbf{UTDE} & \cellcolor{green!20}0.492{\footnotesize$\pm$0.0036} & \cellcolor{green!20}0.740{\footnotesize$\pm$0.0015} & 0.303{\footnotesize$\pm$0.0198} & 0.451{\footnotesize$\pm$0.0120} & 0.823{\footnotesize$\pm$0.0118} & 0.608{\footnotesize$\pm$0.0902} & 0.394{\footnotesize$\pm$0.0066} & 0.193{\footnotesize$\pm$0.0063} & 0.195{\footnotesize$\pm$0.0056}\\
    \textbf{ShaSpec} & 0.481{\footnotesize$\pm$0.0046} & \cellcolor{green!20}0.734{\footnotesize$\pm$0.0016} & 0.341{\footnotesize$\pm$0.0226} & 0.453{\footnotesize$\pm$0.0006} & 0.833{\footnotesize$\pm$0.0035} & 0.580{\footnotesize$\pm$0.0254} & 0.398{\footnotesize$\pm$0.0053} & \bf 0.194{\footnotesize$\pm$0.0082} & \bf 0.198{\footnotesize$\pm$0.0074}\\
    \textbf{Flex-MoE} & \cellcolor{green!20}0.485{\footnotesize$\pm$0.0004} & 0.733{\footnotesize$\pm$0.0019} & 0.312{\footnotesize$\pm$0.0082} & 0.438{\footnotesize$\pm$0.0156} & 0.832{\footnotesize$\pm$0.0059} & 0.516{\footnotesize$\pm$0.0166} & 0.398{\footnotesize$\pm$0.0022} & 0.187{\footnotesize$\pm$0.0056} & 0.195{\footnotesize$\pm$0.0014}\\
    \textbf{DrFuse} & \cellcolor{green!20}0.493{\footnotesize$\pm$0.0024} & \cellcolor{green!20}0.739{\footnotesize$\pm$0.0008} & \cellcolor{green!20}0.376{\footnotesize$\pm$0.0196} & \bf \cellcolor{green!20}0.481{\footnotesize$\pm$0.0090} & 0.838{\footnotesize$\pm$0.0034} & \bf 0.637{\footnotesize$\pm$0.0328} & \bf 0.406{\footnotesize$\pm$0.0067} & 0.185{\footnotesize$\pm$0.0123} & 0.195{\footnotesize$\pm$0.0046}\\
    \textbf{SMIL} & 0.456{\footnotesize$\pm$0.0012} & 0.715{\footnotesize$\pm$0.0020} & 0.293{\footnotesize$\pm$0.0144} & \cellcolor{green!20}0.470{\footnotesize$\pm$0.0122} & 0.834{\footnotesize$\pm$0.0080} & 0.630{\footnotesize$\pm$0.0389} & 0.399{\footnotesize$\pm$0.0027} & 0.146{\footnotesize$\pm$0.0023} & 0.179{\footnotesize$\pm$0.0054}\\
    \textbf{HEALNet} & 0.471{\footnotesize$\pm$0.0009} & 0.726{\footnotesize$\pm$0.0012} & 0.210{\footnotesize$\pm$0.0186} & 0.451{\footnotesize$\pm$0.0074} & 0.836{\footnotesize$\pm$0.0039} & 0.558{\footnotesize$\pm$0.0204} & 0.404{\footnotesize$\pm$0.0013} & 0.187{\footnotesize$\pm$0.0119} & 0.195{\footnotesize$\pm$0.0027}\\
    \textbf{M3Care} & \cellcolor{green!20}0.488{\footnotesize$\pm$0.0026} & \cellcolor{green!20}0.739{\footnotesize$\pm$0.0014} & 0.275{\footnotesize$\pm$0.0251} & 0.445{\footnotesize$\pm$0.0115} & 0.828{\footnotesize$\pm$0.0009} & 0.590{\footnotesize$\pm$0.0243} & 0.400{\footnotesize$\pm$0.0063} & 0.160{\footnotesize$\pm$0.0067} & 0.185{\footnotesize$\pm$0.0077}\\
    \textbf{UMSE} & 0.446{\footnotesize$\pm$0.0008} & 0.711{\footnotesize$\pm$0.0011} & 0.186{\footnotesize$\pm$0.0020} & 0.402{\footnotesize$\pm$0.0133} & 0.802{\footnotesize$\pm$0.0131} & 0.474{\footnotesize$\pm$0.0083} & 0.396{\footnotesize$\pm$0.0058} & 0.157{\footnotesize$\pm$0.0115} & 0.186{\footnotesize$\pm$0.0098}\\
    \textbf{AUG} & \cellcolor{green!20}0.493{\footnotesize$\pm$0.0022} & \bf \cellcolor{green!20}0.742{\footnotesize$\pm$0.0013} & 0.295{\footnotesize$\pm$0.0156} & \cellcolor{green!20}0.474{\footnotesize$\pm$0.0096} & 0.836{\footnotesize$\pm$0.0013} & 0.578{\footnotesize$\pm$0.0492} & 0.401{\footnotesize$\pm$0.0035} & 0.170{\footnotesize$\pm$0.0110} & 0.189{\footnotesize$\pm$0.0097}\\
    \textbf{InfoReg} & \bf \cellcolor{green!20}0.495{\footnotesize$\pm$0.0026} & \cellcolor{green!20}0.741{\footnotesize$\pm$0.0013} & \cellcolor{green!20}0.376{\footnotesize$\pm$0.0054} & \cellcolor{green!20}0.480{\footnotesize$\pm$0.0016} & 0.829{\footnotesize$\pm$0.0065} & 0.635{\footnotesize$\pm$0.0366} & 0.402{\footnotesize$\pm$0.0034} & 0.174{\footnotesize$\pm$0.0168} & 0.188{\footnotesize$\pm$0.0091}\\
    \bottomrule
    \end{tabular}
    }
\end{table}

\begin{table}
\caption{Disease-wise AUPRC for the disease phenotype classification task on the matched cohort. \sethlcolor{pink!50}\hl{Pink-highlighted cells} indicate the best-performing uni-modal method for each disease.
\sethlcolor{green!20}\hl{Green-highlighted cells} denote results that are statistically significantly better than the best uni-modal method based on a permutation test ($p<0.05$).
Boldface values indicate the best-performing model for each disease (row).}
\label{tab:disease_break_down:matched}
\resizebox{\textwidth}{!}{
\setlength{\tabcolsep}{2pt}
  \begin{tabular}{lccccccccccccccccc}
    \toprule
    Disease & Transformer & LSTM & ResNet & LateFusion & MedFuse & MMTM & DAFT & UTDE & ShaSpec & Flex-MoE & DrFuse & SMIL & HEALNet & M3Care & UMSE & AUG & InfoReg\\\midrule
    \textbf{CHF} & \cellcolor{pink!50}0.705 & 0.691 & 0.696 & \cellcolor{green!20}\bf 0.742 & \cellcolor{green!20}0.737 & 0.704 & 0.728 & \cellcolor{green!20}0.736 & \cellcolor{green!20}0.737 & \cellcolor{green!20}0.735 & \cellcolor{green!20}0.735 & \cellcolor{green!20}0.732 & 0.699 & \cellcolor{green!20}0.735 & 0.623 & \cellcolor{green!20}0.741 & \cellcolor{green!20}0.742\\
    \textbf{CorAth/HD} & 0.487 & 0.491 & \cellcolor{pink!50}0.515 & \cellcolor{green!20}0.537 & \cellcolor{green!20}0.544 & 0.543 & \cellcolor{green!20}0.544 & \cellcolor{green!20}0.550 & \cellcolor{green!20}0.548 & \cellcolor{green!20}0.550 & \cellcolor{green!20}\bf 0.553 & 0.519 & 0.497 & \cellcolor{green!20}0.550 & 0.475 & \cellcolor{green!20}0.538 & \cellcolor{green!20}0.550\\
    \textbf{Liver Dis.} & 0.325 & 0.329 & \cellcolor{pink!50}0.334 & \cellcolor{green!20}0.369 & \cellcolor{green!20}0.364 & 0.341 & \cellcolor{green!20}0.367 & \cellcolor{green!20}\bf 0.382 & 0.365 & 0.352 & 0.355 & \cellcolor{green!20}0.359 & 0.331 & \cellcolor{green!20}0.363 & 0.304 & \cellcolor{green!20}0.378 & 0.378\\
    \textbf{COPD} & 0.310 & 0.284 & \cellcolor{pink!50}0.378 & \cellcolor{green!20}0.402 & 0.380 & 0.381 & 0.365 & \cellcolor{green!20}0.385 & 0.376 & 0.372 & \cellcolor{green!20}0.398 & 0.359 & 0.341 & \cellcolor{green!20}0.393 & 0.339 & 0.396 & \cellcolor{green!20}\bf 0.406\\
    \textbf{HTN} & \cellcolor{pink!50}0.543 & 0.534 & 0.510 & 0.561 & 0.558 & 0.537 & \cellcolor{green!20}0.564 & \cellcolor{green!20}\bf 0.577 & \cellcolor{green!20}0.572 & 0.562 & \cellcolor{green!20}0.573 & 0.566 & 0.544 & \cellcolor{green!20}0.568 & 0.558 & 0.561 & 0.561\\
    \textbf{AMI} & \cellcolor{pink!50}0.234 & 0.216 & 0.169 & 0.236 & 0.231 & 0.252 & 0.235 & 0.252 & 0.234 & \cellcolor{green!20}0.246 & \cellcolor{green!20}0.272 & 0.200 & 0.245 & \cellcolor{green!20}\bf 0.278 & \cellcolor{green!20}0.253 & 0.246 & 0.259\\
    \textbf{CD} & 0.508 & 0.395 & \cellcolor{pink!50}0.648 & \cellcolor{green!20}0.687 & 0.683 & 0.593 & 0.686 & \cellcolor{green!20}0.682 & 0.681 & 0.685 & 0.673 & 0.657 & 0.461 & 0.679 & 0.238 & 0.675 & \cellcolor{green!20}\bf 0.694\\
    \textbf{A. CVD} & \cellcolor{pink!50}0.486 & 0.480 & 0.248 & 0.491 & 0.474 & 0.481 & 0.470 & 0.485 & 0.461 & 0.479 & \cellcolor{green!20}\bf 0.514 & 0.412 & \cellcolor{green!20}0.511 & 0.476 & 0.459 & \cellcolor{green!20}0.490 & 0.495\\
    \textbf{GIB} & \cellcolor{pink!50}0.171 & 0.163 & 0.139 & 0.182 & \cellcolor{green!20}\bf 0.183 & 0.170 & 0.165 & \cellcolor{green!20}0.182 & 0.168 & 0.168 & 0.182 & 0.153 & 0.164 & \cellcolor{green!20}0.176 & 0.144 & 0.174 & 0.171\\
    \textbf{DLM} & 0.524 & \cellcolor{pink!50}0.525 & 0.512 & 0.535 & 0.536 & \cellcolor{green!20}0.540 & 0.531 & 0.548 & \cellcolor{green!20}0.540 & 0.529 & 0.546 & 0.530 & 0.536 & 0.529 & 0.527 & \bf 0.550 & 0.528\\
    \textbf{PNA} & \cellcolor{pink!50}0.470 & 0.455 & 0.412 & 0.493 & 0.475 & 0.465 & 0.480 & 0.495 & 0.471 & 0.487 & \cellcolor{green!20}\bf 0.500 & 0.467 & 0.478 & 0.493 & 0.455 & 0.486 & 0.493\\
    \textbf{HTC} & \cellcolor{pink!50}0.567 & 0.529 & 0.395 & 0.557 & 0.552 & 0.505 & 0.558 & 0.566 & \cellcolor{green!20}\bf 0.584 & 0.572 & 0.578 & 0.522 & 0.579 & 0.565 & 0.583 & 0.570 & 0.561\\
    \textbf{DMC} & \cellcolor{pink!50}0.601 & 0.537 & 0.215 & 0.574 & 0.525 & 0.580 & 0.564 & 0.590 & 0.551 & 0.566 & 0.589 & 0.477 & 0.600 & 0.578 & \cellcolor{green!20}\bf 0.608 & 0.564 & 0.586\\
    \textbf{URD} & \cellcolor{pink!50}0.220 & 0.167 & 0.164 & 0.208 & 0.176 & 0.193 & 0.185 & 0.204 & 0.172 & 0.183 & 0.194 & 0.129 & 0.207 & 0.183 & 0.161 & \cellcolor{green!20}\bf 0.233 & 0.228\\
    \textbf{Sepsis} & \cellcolor{pink!50}0.566 & 0.553 & 0.417 & 0.556 & 0.544 & 0.550 & 0.537 & 0.570 & 0.556 & 0.569 & \cellcolor{green!20}\bf 0.576 & 0.535 & 0.544 & 0.567 & 0.545 & 0.571 & 0.574\\
    \textbf{RF} & \cellcolor{pink!50}0.664 & 0.648 & 0.587 & 0.665 & 0.653 & 0.650 & 0.658 & 0.663 & 0.660 & 0.663 & 0.663 & 0.655 & 0.656 & 0.662 & 0.651 & \bf 0.666 & 0.664\\
    \textbf{PTX} & 0.177 & 0.160 & \cellcolor{pink!50}0.208 & 0.204 & 0.195 & 0.178 & 0.197 & 0.202 & 0.174 & 0.189 & 0.200 & 0.193 & 0.171 & 0.196 & 0.166 & 0.204 & \bf 0.230\\
    \textbf{LRD} & 0.205 & \cellcolor{pink!50}\bf 0.219 & 0.198 & 0.210 & 0.217 & 0.212 & 0.206 & 0.209 & 0.217 & 0.194 & 0.205 & 0.204 & 0.209 & 0.210 & 0.207 & 0.212 & 0.209\\
    \textbf{ARF} & \cellcolor{pink!50}0.656 & 0.642 & 0.517 & 0.650 & 0.639 & 0.646 & 0.641 & 0.658 & 0.644 & 0.654 & 0.648 & 0.591 & 0.658 & 0.648 & \bf 0.677 & 0.658 & 0.657\\
    \textbf{DM-nC} & \cellcolor{pink!50}\bf 0.388 & 0.350 & 0.269 & 0.366 & 0.334 & 0.384 & 0.356 & 0.382 & 0.350 & 0.367 & 0.373 & 0.294 & 0.379 & 0.373 & 0.385 & 0.383 & 0.378\\
    \textbf{SurgCmp} & \cellcolor{pink!50}0.377 & 0.375 & 0.345 & 0.379 & 0.374 & 0.373 & 0.368 & 0.378 & \bf 0.387 & 0.374 & 0.381 & 0.354 & 0.369 & 0.374 & 0.382 & 0.382 & 0.382\\
    \textbf{CKD} & \cellcolor{pink!50}0.601 & 0.570 & 0.427 & 0.598 & 0.595 & 0.544 & 0.602 & 0.609 & 0.613 & 0.612 & \bf 0.616 & 0.561 & 0.608 & 0.606 & 0.608 & 0.612 & 0.605\\
    \textbf{Arrhy} & \cellcolor{pink!50}0.744 & 0.732 & 0.606 & \bf 0.746 & 0.722 & 0.738 & 0.732 & 0.737 & 0.726 & 0.735 & 0.731 & 0.672 & 0.730 & 0.735 & 0.555 & 0.746 & 0.744\\
    \textbf{FluEle} & \cellcolor{pink!50}0.691 & 0.686 & 0.667 & 0.702 & 0.695 & 0.689 & 0.697 & 0.691 & 0.690 & 0.694 & 0.689 & 0.688 & 0.698 & 0.701 & 0.692 & \bf 0.709 & 0.702\\
    \textbf{Shock} & \cellcolor{pink!50}0.575 & 0.566 & 0.418 & 0.575 & 0.555 & 0.561 & 0.560 & 0.568 & 0.555 & 0.576 & \bf 0.578 & 0.561 & 0.572 & 0.565 & 0.560 & 0.573 & 0.574\\
\midrule
Avg.Rank & 9.4 & 13.3 & 15.3 & 6.0 & 10.2 & 10.2 & 10.1 & 5.4 & 8.8 & 8.0 & 5.4 & 13.4 & 9.5 & 7.3 & 11.3 & 4.8 & 4.6\\
\bottomrule
  \end{tabular}
}

\begin{tablenotes}[flushleft]\footnotesize
\item \textit{Disease abbreviations:}
\textbf{CHF}, congestive heart failure; nonhypertensive;
\textbf{CorAth/HD}, coronary atherosclerosis and other heart disease;
\textbf{Liver Dis.}, other liver diseases;
\textbf{COPD}, chronic obstructive pulmonary disease and bronchiectasis;
\textbf{HTN}, essential hypertension;
\textbf{AMI}, acute myocardial infarction;
\textbf{CD}, conduction disorders;
\textbf{A. CVD}, acute cerebrovascular disease;
\textbf{GIB}, gastrointestinal hemorrhage;
\textbf{DLM}, disorders of lipid metabolism;
\textbf{PNA}, pneumonia (except that caused by tuberculosis or sexually transmitted disease);
\textbf{HTC}, hypertension with complications and secondary hypertension;
\textbf{DMC}, diabetes mellitus with complications;
\textbf{URD}, other upper respiratory disease;
\textbf{Sepsis}, septicemia (except in labor);
\textbf{RF}, respiratory failure; insufficiency; arrest (adult);
\textbf{PTX}, pleurisy; pneumothorax; pulmonary collapse;
\textbf{LRD}, other lower respiratory disease;
\textbf{ARF}, acute and unspecified renal failure;
\textbf{DM-nC}, diabetes mellitus without complication;
\textbf{SurgCmp}, complications of surgical procedures or medical care;
\textbf{CKD}, chronic kidney disease;
\textbf{Arrhy}, cardiac dysrhythmias;
\textbf{FluEle}, fluid and electrolyte disorders.

\end{tablenotes}
\end{table}

\subsection{Research Questions}
In this section, we present our benchmarking results and analyses, organized around the following research questions: 

\textbf{RQ1: When does fusing multimodal data benefit medical tasks?}  
While integrating more modalities is generally expected to provide additional information, it remains unclear whether multimodal fusion consistently improves performance, and under what conditions such gains emerge. We evaluate unimodal and multimodal models across various clinical tasks and conduct statistical tests to assess whether multimodal fusion yields significant improvements over the strongest unimodal baseline.

\textbf{RQ2: How do different fusion strategies compare under the modality-complete settings?}  When all modalities are available, fusion models are expected to jointly exploit complementary information. Existing approaches employ different strategies to aggregate informationa cross modalities. We systematically compare those fusion strategies across tasks and evaluation metrics to identify what design factors most effectively support multimodal learning.

\textbf{RQ3: How robust are multimodal fusion methods to modality missingness?}  
Modality missingness is common in clinical practice due to heterogeneous diagnostic workflows. While some methods explicitly model missing modalities, others assume complete inputs. We benchmark both categories under realistic missingness settings, applying zero imputation to models not designed for incomplete inputs, to examine their robustness.

\textbf{RQ4: Do multimodal fusion methods achieve algorithmic fairness?}  
Beyond predictive accuracy, understanding fairness is essential for clinical deployment, as model behavior may vary across sensitive attributes. We evaluate multimodal models using fairness metrics: AUPRC gap, TPR gap, FPR gap, and ECE gap, to characterize performance disparities across racial subgroups.

\subsection{Benefits of Multimodal Fusion (RQ1)}
We first benchmark all models on a matched cohort with complete modalities to assess when multimodal fusion provides measurable improvements over strong unimodal baselines, with aggregate task-level results reported in \Cref{tab:prediction_matched} and disease-wise phenotyping performance in \Cref{tab:disease_break_down:matched}.

\textbf{\textit{Finding 1: Multimodal fusion generally improves performance under complete-modality settings.}} 
On the matched subset, multimodal models consistently achieve the best performance across all tasks and evaluation metrics (see \Cref{tab:prediction_matched}). For phenotyping, leading fusion approaches such as InfoReg (AUPRC 0.495) and AUG (AUROC 0.742) significantly outperform the strongest unimodal baseline (EHR Transformer: AUPRC 0.472, AUROC 0.724). Similar trends are observed for mortality prediction, where DrFuse achieves the strongest AUPRC (0.481) compared to the LSTM (0.451), and MedFuse attains the highest AUROC (0.851 vs. 0.821 from Transformer).  For LoS prediction, multimodal models provide more modest gains, with ShaSpec achieving the highest Kappa (0.198) relative to the best unimodal model (LSTM, 0.192). These results demonstrate that when modalities are complete, chest radiographs provide complementary information to structured EHR data, enabling richer patient representations and improved predictive performance.

To quantify these improvements, we further conduct permutation tests comparing each multimodal model against the best unimodal baseline. While statistically significant gains are frequently observed for phenotyping, improvements for mortality and LoS are generally smaller and often not significant. 
A plausible explanation is that these endpoints are already largely explained by temporally dense physiological signals and laboratory measurements in EHR, while CXR provides a more indirect and condition-dependent signal, leading to less consistent marginal gains.

\textbf{\textit{Finding 2: Multimodal benefits concentrate in modality-distributed phenotypes.}} 
Building on the aggregate results, phenotyping exhibits the largest and most stable gains from multimodal fusion, motivating a closer disease-level analysis~(\Cref{tab:disease_break_down:matched}). Conditions such as congestive heart failure, coronary heart disease, COPD, and liver disease benefit most from multimodal learning. These diseases are characterized by modality-distributed phenotypes, where CXR capture structural or pulmonary manifestations while EHR encodes longitudinal risk factors, laboratory measurements, and comorbidity history. Multimodal models therefore leverage complementary signals that are inaccessible to either modality alone.

In contrast, conditions such as sepsis are primarily driven by acute physiological changes and laboratory abnormalities, with chest radiographs offering only indirect or nonspecific cues, resulting in more limited fusion benefits. This distinction between modality-distributed and modality-localized phenotypes provides a clinical explanation for the heterogeneous performance gains observed across disease categories.

\subsection{Fusion Strategy Comparison (RQ2)}
To further answer the question of which fusion strategies are most effective, and better understand the underlying reasons, we compare 14 fusion methods spanning diverse architectural paradigms, from simple concatenation to sophisticated cross-modal attention, representation disentanglement, and training-time regularization.

\textbf{\textit{Finding 3: Cross-modal learning mechanisms capture clinically meaningful dependencies that concatenation misses.}}
\Cref{tab:prediction_matched} demonstrates that methods enabling cross-modal learning substantially outperform naive late fusion, revealing that clinically relevant interactions between EHR and CXR cannot be captured by treating them as independent information sources. For phenotyping, methods that facilitate cross-modal information exchange, such as InfoReg (0.495 AUPRC), AUG (0.493 AUPRC), DrFuse (0.493 AUPRC), and UTDE (0.492 AUPRC), all surpass late fusion (0.489 AUPRC). This performance gap reflects the clinical reality that imaging findings should be interpreted in the context of physiological state and vice versa. Pneumonia (PNA) in \cref{tab:disease_break_down:matched} exemplifies this phenomenon: neither unimodal baseline alone is sufficient (EHR Transformer: 0.470, CXR ResNet: 0.412), and while late fusion improves performance to 0.493 by combining both information sources, advanced cross-modal learning methods achieve further gains (DrFuse: 0.500, UTDE: 0.495). This improvement directly reflects the clinical diagnostic process: pulmonary infiltrates visible on CXR have different diagnostic implications depending on whether the patient exhibits fever, leukocytosis, and hypoxemia (suggesting bacterial pneumonia) versus absence of systemic inflammatory response (suggesting atelectasis or chronic changes). Late fusion, which processes each modality independently before combining predictions, cannot capture these conditional dependencies. For mortality prediction, DrFuse (0.481 AUPRC) and InfoReg (0.480 AUPRC) also substantially outperform late fusion (0.433 AUPRC). The disease-wise phenotype classification analysis (\cref{tab:disease_break_down:matched}) shows that AUG (rank 4.8), InfoReg (4.6), UTDE (5.4), and DrFuse (5.4) consistently outperform late fusion (6.0). Nevertheless, late fusion achieves the best performance in tasks like CHF and Arrhy classification, suggesting that the value of cross-modal learning also depends on the reasoning complexity when integrating EHR and imaging evidence.

\begin{table}\small
\centering
\caption{Predictive Performance over the base cohort. \sethlcolor{pink!50}\hl{Pink-highlighted cells} indicate the best uni-modal method for each disease.
\sethlcolor{green!20}\hl{Green-highlighted cells} denote results that are statistically significantly better than the best uni-modal method based on a permutation test ($p<0.05$).
Boldface values indicate the best-performing model for each evaluation metric (column).}
\label{tab:prediction_full}
\setlength{\tabcolsep}{3pt}
\resizebox{\textwidth}{!}{
\begin{tabular}{lcccccccccc}
    \toprule
     & \multicolumn{3}{c}{\textbf{Phenotyping}} & \multicolumn{3}{c}{\textbf{Mortality}} & \multicolumn{3}{c}{\textbf{Length of Stay}} \\\cmidrule(lr){2-4}\cmidrule(lr){5-7}\cmidrule(lr){8-10}
     Models & AUPRC & AUROC & F1 & AUPRC & AUROC & F1 & ACC & F1 & Kappa \\
    \midrule
    \textbf{Transformer} & \cellcolor{pink!50}0.479{\footnotesize$\pm$0.0011} & \cellcolor{pink!50}0.759{\footnotesize$\pm$0.0003} & \cellcolor{pink!50}0.362{\footnotesize$\pm$0.0041} & \cellcolor{pink!50}0.504{\footnotesize$\pm$0.0059} & \cellcolor{pink!50}0.867{\footnotesize$\pm$0.0010} & \cellcolor{pink!50}\bf 0.679{\footnotesize$\pm$0.0133} & \cellcolor{pink!50}0.417{\footnotesize$\pm$0.0033} & \cellcolor{pink!50}0.198{\footnotesize$\pm$0.0144} & \cellcolor{pink!50}0.204{\footnotesize$\pm$0.0053}\\
    \textbf{LSTM} & 0.468{\footnotesize$\pm$0.0013} & 0.755{\footnotesize$\pm$0.0003} & 0.334{\footnotesize$\pm$0.0224} & 0.480{\footnotesize$\pm$0.0131} & 0.861{\footnotesize$\pm$0.0061} & 0.661{\footnotesize$\pm$0.0232} & 0.412{\footnotesize$\pm$0.0034} & 0.178{\footnotesize$\pm$0.0043} & 0.197{\footnotesize$\pm$0.0034}\\
    \textbf{ResNet} & 0.280{\footnotesize$\pm$0.0021} & 0.565{\footnotesize$\pm$0.0001} & 0.115{\footnotesize$\pm$0.0104} & 0.134{\footnotesize$\pm$0.0086} & 0.565{\footnotesize$\pm$0.0143} & 0.476{\footnotesize$\pm$0.0009} & 0.354{\footnotesize$\pm$0.0005} & 0.106{\footnotesize$\pm$0.0008} & 0.041{\footnotesize$\pm$0.0018}\\\midrule
    \textbf{LateFusion} & 0.485{\footnotesize$\pm$0.0006} & 0.762{\footnotesize$\pm$0.0017} & 0.289{\footnotesize$\pm$0.0163} & 0.500{\footnotesize$\pm$0.0067} & 0.867{\footnotesize$\pm$0.0023} & 0.647{\footnotesize$\pm$0.0193} & 0.416{\footnotesize$\pm$0.0011} & \cellcolor{green!20}0.200{\footnotesize$\pm$0.0096} & \cellcolor{green!20}0.210{\footnotesize$\pm$0.0046}\\
    \textbf{MedFuse} & 0.469{\footnotesize$\pm$0.0016} & 0.756{\footnotesize$\pm$0.0009} & 0.220{\footnotesize$\pm$0.0003} & 0.508{\footnotesize$\pm$0.0069} & \bf 0.874{\footnotesize$\pm$0.0025} & 0.636{\footnotesize$\pm$0.0263} & \cellcolor{green!20}0.418{\footnotesize$\pm$0.0028} & \bf \cellcolor{green!20}0.203{\footnotesize$\pm$0.0085} & \bf \cellcolor{green!20}0.213{\footnotesize$\pm$0.0028}\\
    \textbf{MMTM} & 0.471{\footnotesize$\pm$0.0009} & 0.755{\footnotesize$\pm$0.0003} & 0.339{\footnotesize$\pm$0.0145} & 0.475{\footnotesize$\pm$0.0031} & 0.865{\footnotesize$\pm$0.0004} & 0.663{\footnotesize$\pm$0.0209} & 0.404{\footnotesize$\pm$0.0020} & 0.162{\footnotesize$\pm$0.0118} & 0.171{\footnotesize$\pm$0.0032}\\
    \textbf{DAFT} & 0.475{\footnotesize$\pm$0.0025} & 0.756{\footnotesize$\pm$0.0007} & 0.235{\footnotesize$\pm$0.0192} & 0.495{\footnotesize$\pm$0.0057} & 0.870{\footnotesize$\pm$0.0019} & 0.592{\footnotesize$\pm$0.0401} & 0.418{\footnotesize$\pm$0.0007} & \cellcolor{green!20}0.198{\footnotesize$\pm$0.0106} & 0.208{\footnotesize$\pm$0.0031}\\
    \textbf{UTDE} & \cellcolor{green!20}0.485{\footnotesize$\pm$0.0011} & 0.764{\footnotesize$\pm$0.0004} & 0.279{\footnotesize$\pm$0.0105} & 0.503{\footnotesize$\pm$0.0048} & 0.868{\footnotesize$\pm$0.0015} & 0.621{\footnotesize$\pm$0.0308} & 0.416{\footnotesize$\pm$0.0006} & \cellcolor{green!20}0.201{\footnotesize$\pm$0.0057} & 0.207{\footnotesize$\pm$0.0002}\\
    \textbf{ShaSpec} & \cellcolor{green!20}0.485{\footnotesize$\pm$0.0009} & \cellcolor{green!20}0.763{\footnotesize$\pm$0.0003} & 0.347{\footnotesize$\pm$0.0139} & 0.500{\footnotesize$\pm$0.0068} & 0.869{\footnotesize$\pm$0.0027} & 0.671{\footnotesize$\pm$0.0215} & \cellcolor{green!20}0.421{\footnotesize$\pm$0.0008} & 0.191{\footnotesize$\pm$0.0036} & 0.209{\footnotesize$\pm$0.0013}\\
    \textbf{Flex-MoE} & 0.484{\footnotesize$\pm$0.0023} & 0.762{\footnotesize$\pm$0.0016} & 0.316{\footnotesize$\pm$0.0084} & 0.507{\footnotesize$\pm$0.0061} & 0.868{\footnotesize$\pm$0.0014} & 0.630{\footnotesize$\pm$0.0524} & 0.414{\footnotesize$\pm$0.0025} & 0.188{\footnotesize$\pm$0.0092} & 0.206{\footnotesize$\pm$0.0038}\\
    \textbf{DrFuse} & 0.484{\footnotesize$\pm$0.0004} & \cellcolor{green!20}0.764{\footnotesize$\pm$0.0005} & 0.361{\footnotesize$\pm$0.0048} & 0.500{\footnotesize$\pm$0.0078} & 0.874{\footnotesize$\pm$0.0035} & 0.656{\footnotesize$\pm$0.0086} & 0.419{\footnotesize$\pm$0.0015} & 0.191{\footnotesize$\pm$0.0148} & 0.196{\footnotesize$\pm$0.0051}\\
    \textbf{SMIL} & 0.447{\footnotesize$\pm$0.0004} & 0.742{\footnotesize$\pm$0.0007} & 0.296{\footnotesize$\pm$0.0063} & 0.478{\footnotesize$\pm$0.0164} & 0.860{\footnotesize$\pm$0.0089} & 0.631{\footnotesize$\pm$0.0231} & 0.417{\footnotesize$\pm$0.0020} & 0.157{\footnotesize$\pm$0.0059} & 0.192{\footnotesize$\pm$0.0051}\\
    \textbf{HEALNet} & 0.475{\footnotesize$\pm$0.0008} & 0.758{\footnotesize$\pm$0.0005} & 0.220{\footnotesize$\pm$0.0041} & 0.491{\footnotesize$\pm$0.0055} & 0.873{\footnotesize$\pm$0.0018} & 0.623{\footnotesize$\pm$0.0421} & 0.418{\footnotesize$\pm$0.0011} & 0.191{\footnotesize$\pm$0.0117} & 0.200{\footnotesize$\pm$0.0062}\\
    \textbf{M3Care} & \bf \cellcolor{green!20}0.488{\footnotesize$\pm$0.0006} & 0.764{\footnotesize$\pm$0.0007} & 0.271{\footnotesize$\pm$0.0109} & 0.499{\footnotesize$\pm$0.0061} & 0.869{\footnotesize$\pm$0.0011} & 0.624{\footnotesize$\pm$0.0221} & 0.420{\footnotesize$\pm$0.0016} & 0.188{\footnotesize$\pm$0.0110} & 0.204{\footnotesize$\pm$0.0031}\\
    \textbf{UMSE} & 0.455{\footnotesize$\pm$0.0011} & 0.747{\footnotesize$\pm$0.0014} & 0.185{\footnotesize$\pm$0.0110} & 0.438{\footnotesize$\pm$0.0136} & 0.842{\footnotesize$\pm$0.0062} & 0.543{\footnotesize$\pm$0.0244} & 0.406{\footnotesize$\pm$0.0011} & 0.175{\footnotesize$\pm$0.0072} & 0.179{\footnotesize$\pm$0.0044}\\
    \textbf{AUG} & \cellcolor{green!20}0.487{\footnotesize$\pm$0.0008} & \cellcolor{green!20}0.764{\footnotesize$\pm$0.0001} & \cellcolor{green!20}0.364{\footnotesize$\pm$0.0078} & 0.503{\footnotesize$\pm$0.0080} & 0.870{\footnotesize$\pm$0.0014} & 0.579{\footnotesize$\pm$0.0166} & \bf 0.424{\footnotesize$\pm$0.0016} & 0.177{\footnotesize$\pm$0.0060} & 0.205{\footnotesize$\pm$0.0016}\\
    \textbf{InfoReg} & \cellcolor{green!20}0.488{\footnotesize$\pm$0.0004} & \bf \cellcolor{green!20}0.764{\footnotesize$\pm$0.0002} & \bf \cellcolor{green!20}0.386{\footnotesize$\pm$0.0023} & \bf \cellcolor{green!20}0.511{\footnotesize$\pm$0.0016} & 0.871{\footnotesize$\pm$0.0020} & 0.663{\footnotesize$\pm$0.0309} & \cellcolor{green!20}0.421{\footnotesize$\pm$0.0011} & \cellcolor{green!20}0.202{\footnotesize$\pm$0.0045} & 0.210{\footnotesize$\pm$0.0033}\\
        \bottomrule 
    \end{tabular}
}
\end{table}

\begin{figure}
    \centering
    \begin{subfigure}[t]{0.32\textwidth}
        \centering
        \includegraphics[width=\linewidth]{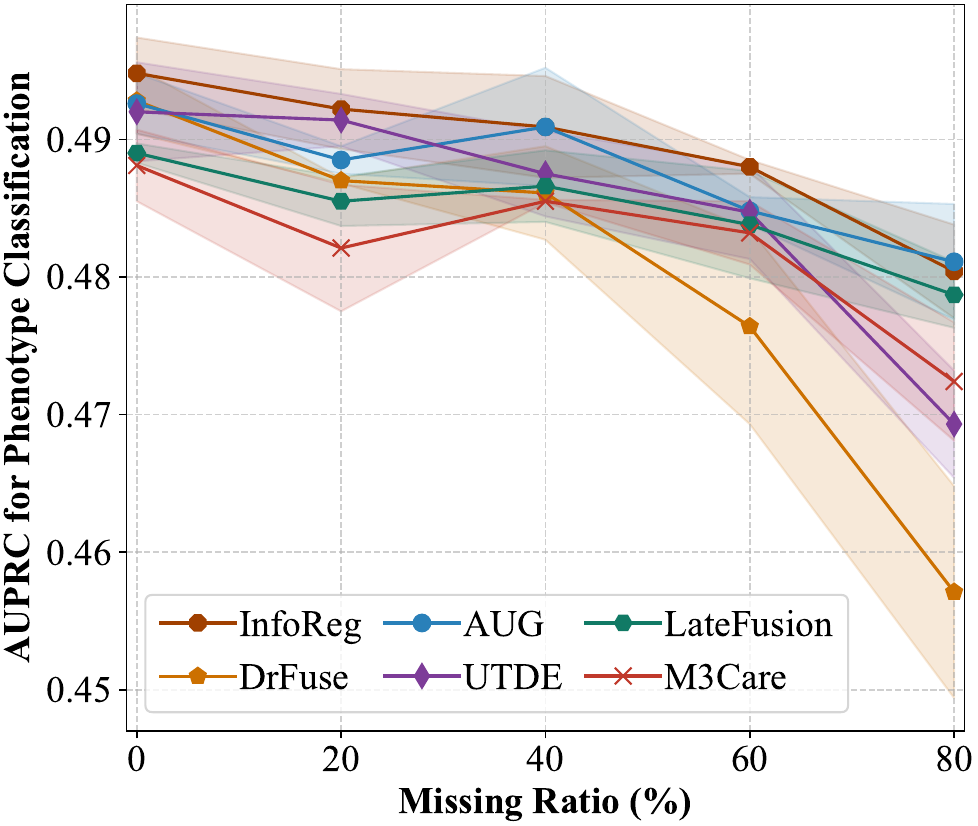}
        \caption{Overall Phenotype Classification}\label{fig:robustness:phenotyping}
    \end{subfigure}
    \begin{subfigure}[t]{0.32\textwidth}
        \centering
        \includegraphics[width=\linewidth]{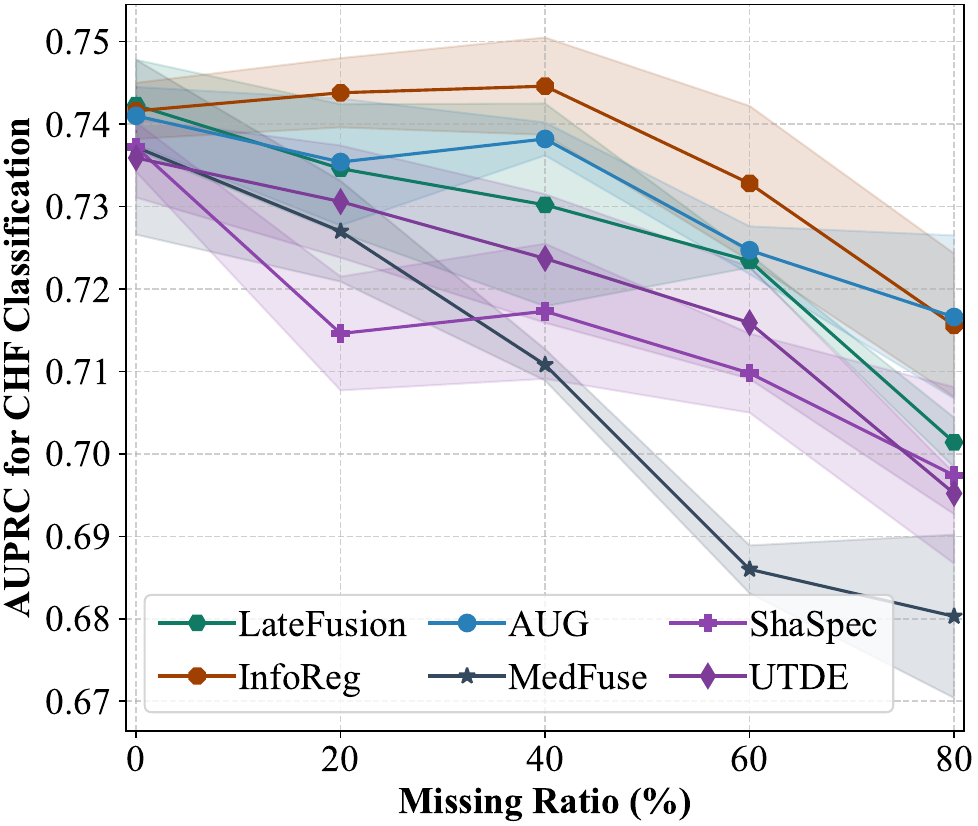}
        \caption{CHF Disease Classification}\label{fig:robustness:chf}
    \end{subfigure}
    \begin{subfigure}[t]{0.32\textwidth}
        \centering
        \includegraphics[width=\linewidth]{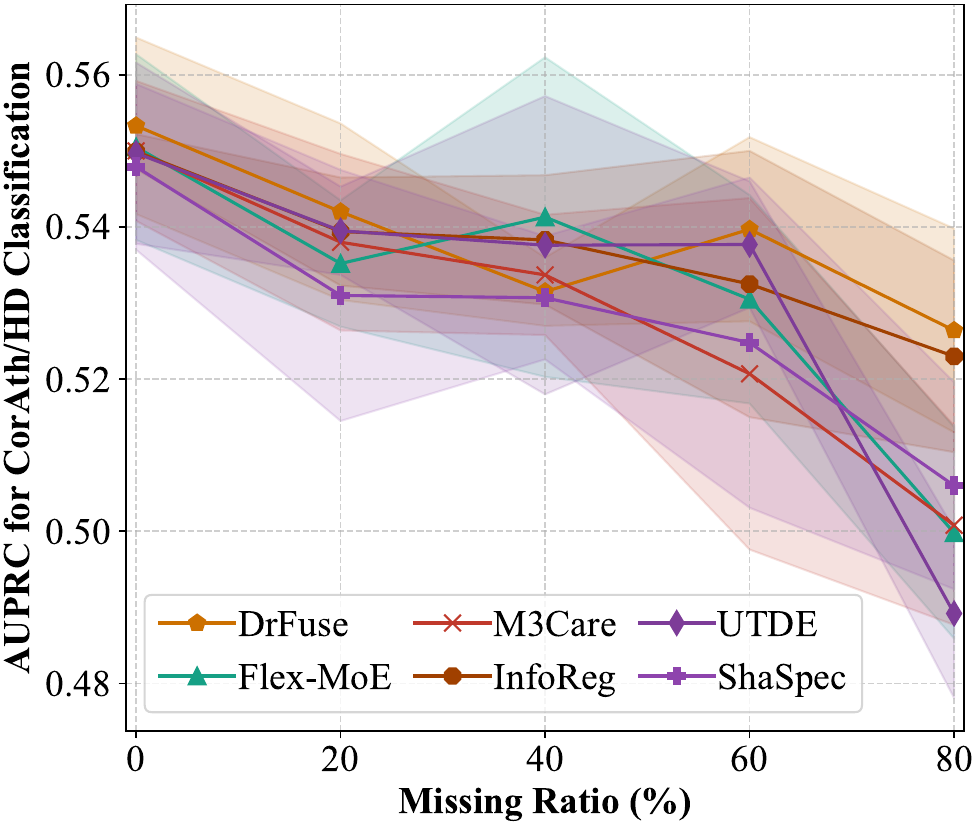}
        \caption{CorAth/HD Disease Classification}\label{fig:robustness:corath}
    \end{subfigure}\\
    \begin{subfigure}[t]{0.32\textwidth}
        \centering
        \includegraphics[width=\linewidth]{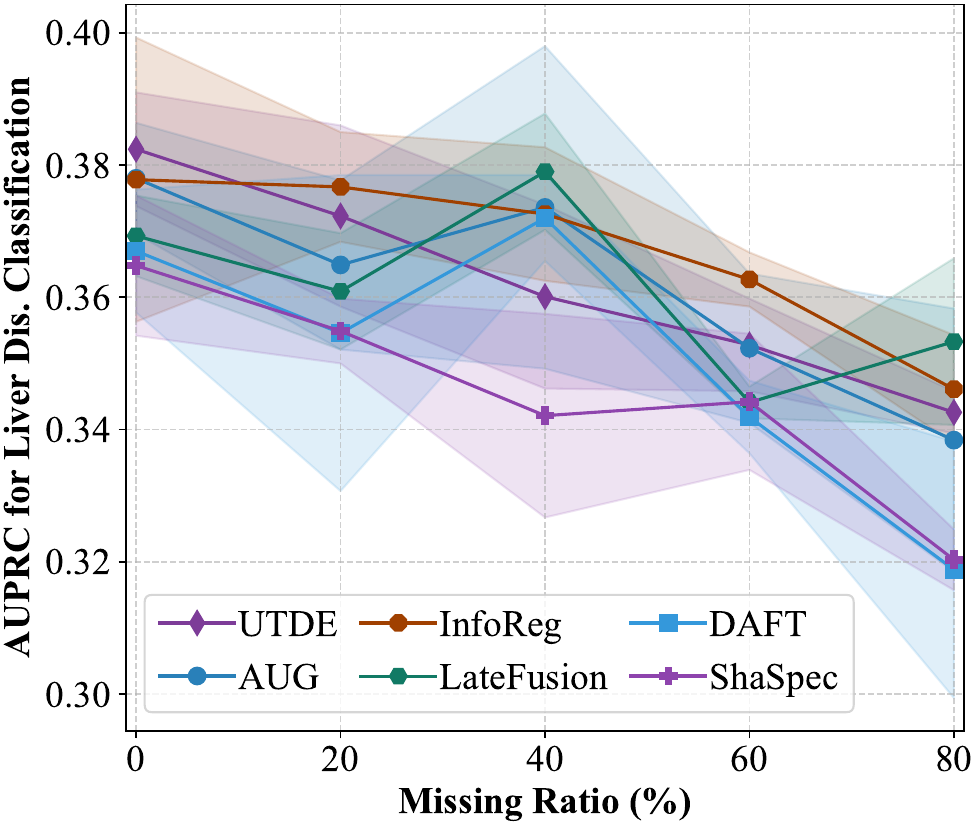}
        \caption{Liver Diseases Classification}\label{fig:robustness:liver}
    \end{subfigure}
    \begin{subfigure}[t]{0.32\textwidth}
        \centering
        \includegraphics[width=\linewidth]{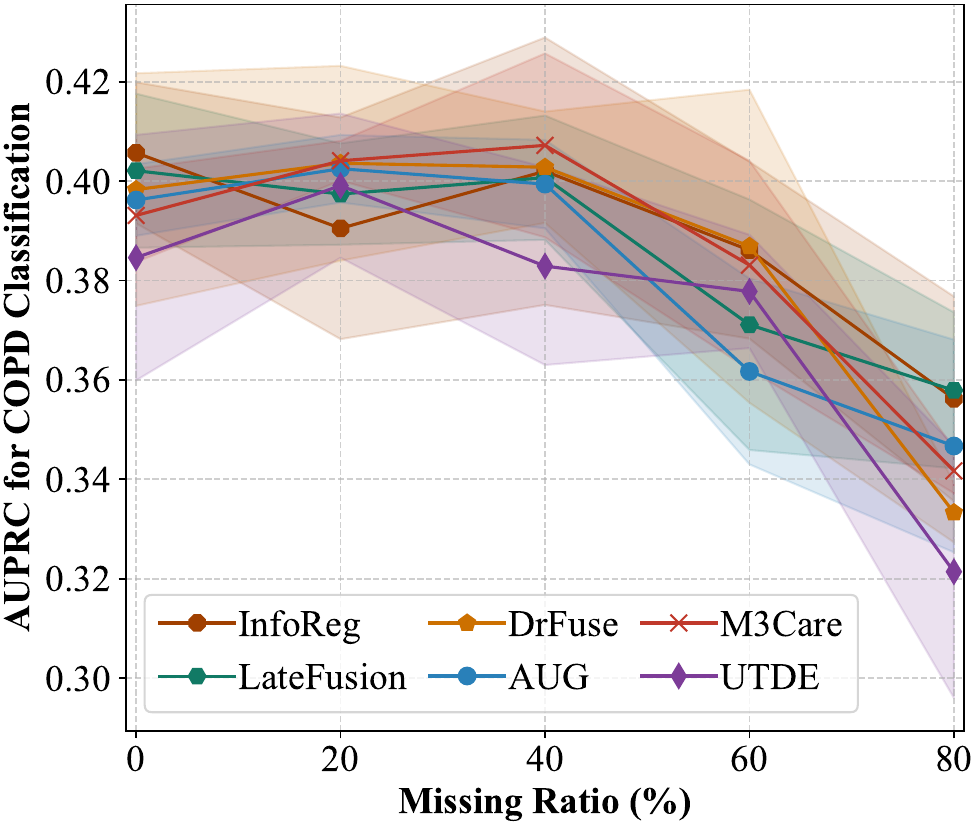}
        \caption{COPD Disease Classification}\label{fig:robustness:chf}
    \end{subfigure}
    \begin{subfigure}[t]{0.32\textwidth}
        \centering
        \includegraphics[width=\linewidth]{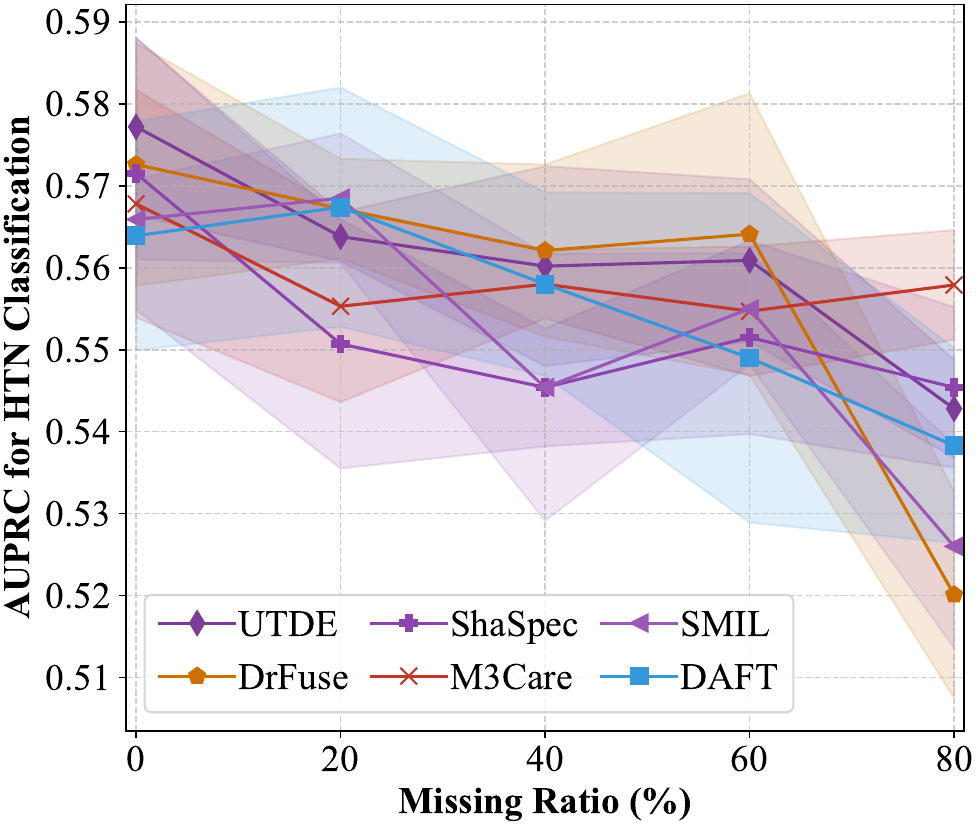}
        \caption{HTN Disease Classification}\label{fig:robustness:corath}
    \end{subfigure}
    \caption{Robustness of the top six performing methods for the overall phenotype classification task and five disease classification tasks under varying ratios of modality missingness. The shaded areas represent standard deviations.}
    \label{fig:robustness}
\end{figure}

\textbf{\textit{Finding 4: EHR's longitudinal richness creates modality imbalance that architectural complexity alone cannot overcome.}}
A striking pattern emerges when examining the top performers: methods that explicitly address modality imbalance (AUG, InfoReg) consistently achieve superior performance despite using relatively simple fusion architectures, ranking 4.8 and 4.6 respectively, across 25 diseases (\cref{tab:disease_break_down:matched}). This superior performance reflects a fundamental characteristic of clinical data: EHR contains rich longitudinal signals, such as hourly vital signs, laboratory trends, medication responses, and physiological trajectories spanning the entire observation window. Such rich data capture disease progression and patient deterioration in granular detail. In contrast, a single frontal chest X-ray, while diagnostically valuable, provides only a static snapshot at one timepoint and is inherently limited in depicting temporal dynamics such as sepsis progression, fluid balance changes, or response to interventions \citep{liu2025multimodal}. This creates a natural information asymmetry where the EHR modality dominates the learning process. The success of InfoReg, which explicitly slows down learning from information-rich modalities during critical training phases, and AUG, which iteratively boosts the classification ability of weaker modalities, demonstrates that addressing this clinical data imbalance is more critical than architectural sophistication. DrFuse (rank 5.4), which is designed for handling missing modalities through disentanglement, also performs competitively. This is likely because separating shared versus modality-specific features prevents the longitudinally rich EHR from overshadowing the complementary but weaker CXR signals. This finding suggests that effective clinical multimodal fusion must account for the inherent information density differences between continuous monitoring data and intermittent imaging studies.

\begin{figure}
    \centering
        \includegraphics[width=\textwidth]{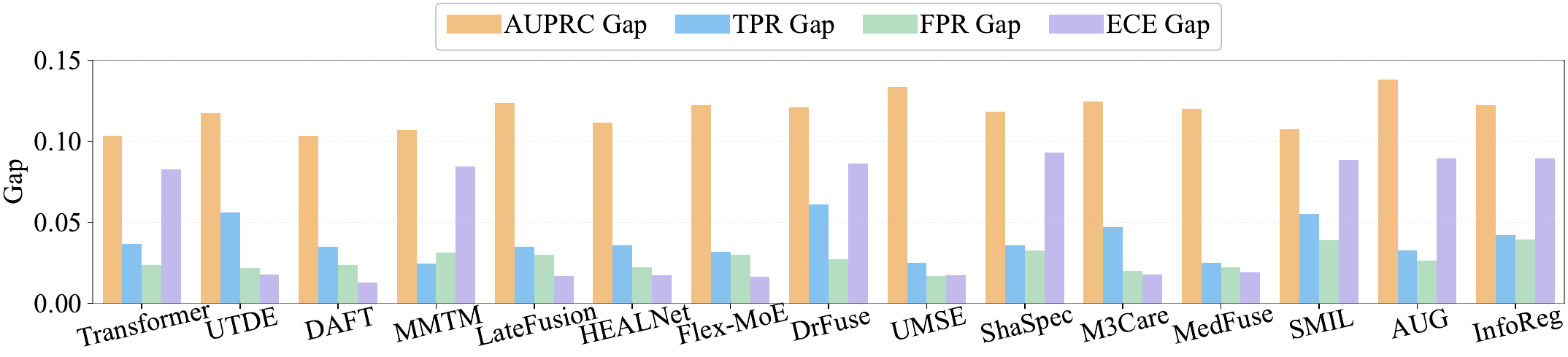}
    \caption{Race Group fairness gaps on the phenotype matched subset.}
    \label{fig:fairness-Race-gaps}
\end{figure}

\subsection{Robustness to Modality Missingness (RQ3)}
In clinical practice, modality missingness is pervasive: approximately 75\% of ICU stays in our base cohort lack paired chest X-rays. We first evaluate all models on this realistic base cohort (\Cref{tab:prediction_full}). To better understand the robustness to modality missing ratios, we further conduct controlled experiments systematically varying CXR missingness from 0\% to 80\% (\Cref{fig:robustness}). For methods not explicitly designed to handle missing modalities, we apply zero-imputation to absent CXR features.

\textbf{\textit{Finding 5: Specialized architectural and algorithmic designs are essential for leveraging incomplete multimodal data.}}
In the base cohort, the EHR-only Transformer establishes a strong benchmark that many multimodal models fail to surpass, for example, the highest mortality F1-score (0.679), and a LoS Kappa score (0.204), outperforming several fusion methods designed for complete data, such as MMTM (0.663 and 0.1714, for mortality F1 and LoS Kappa, respectively). This suggests that naively applying models designed for complete-case scenarios does not guarantee a benefit and often fails once missingness is introduced. On the other hand, architectures specifically designed to handle modality absence are essential to make a better use of the multimodal data. The models explicitly tailored for this challenge, such as MedFuse, M3Care, and HEALNet, outperform both the unimodal baseline and the complete-case fusion methods in most cases. For mortality prediction, MedFuse attains the highest AUROC (0.874). In LoS prediction, MedFuse also leads with the best Kappa (0.213) and F1-score (0.203), surpassing the strong EHR Transformer. 

\textbf{\textit{Finding 6: Severe missingness amplifies modality imbalance, hindering multimodal fusion in realistic settings.}} In the base cohort, EHR is present for all 26,947 ICU stays, providing dense temporal signals spanning the full 48-hour observation window, while CXR is available for only 7,149 cases ($\sim$25\%). This distributional imbalance compounds the inherent information asymmetry (EHR's longitudinal richness vs. CXR's single timepoint), creating an amplified dominance effect during training. Most multimodal fusion methods implicitly assume roughly balanced modality contributions; thus, they optimize primarily on EHR-driven gradients since the CXR signal appears in only a quarter of training samples. The resulting representations encode predominantly EHR patterns, with CXR features remaining underdeveloped even when present. \Cref{tab:prediction_full} shows that InfoReg and AUG outperform most methods in the base cohort. This is likely because they explicitly counteract the amplified modality imbalance in their learning algorithms. Therefore, under the senarios of severe modality missingness, modality imbalance should also be considered when addressing misisng modality problems. 

The controlled missingness experiments in \Cref{fig:robustness} further demonstrates this. For overall phenotyping (\Cref{fig:robustness:phenotyping}), InfoReg and AUG degrade notably slower than DrFuse and M3Care, as the missing ratio increase from 0\% to 80\% missingness. Although DrFuse and M3Care are explicitly designed for handling missing modality, they do not account for the modality imabalnce. Similar patterns can also be observed in disease-specific results as visualzied in \Cref{fig:robustness}. These consistently slower degradation rates demonstrate that modality balancing mechanisms enable graceful degradation by ensuring CXR learns discriminative features during training despite its scarcity.

\subsection{Fairness Across Subgroups (RQ4)}
To assess fairness, we stratify model performance across the attribute of race. We use AUPRC, TPR, FPR, and ECE as the primary metrics and quantify disparities with each metric (difference between the best- and worst-performing subgroups), reporting results on the matched subset with phenotyping task in \cref{fig:fairness-Race-gaps}.

\textbf{\textit{Finding 7: Multimodal fusion does not inherently improve algorithmic fairness.}}
Despite improving predictive performance (Finding 1), multimodal fusion does not consistently reduce subgroup disparities across fairness metrics. As shown in~\Cref{fig:fairness-Race-gaps}, all multimodal models exhibit larger AUPRC gaps than the unimodal Transformer baseline, indicating increased performance disparity across race subgroups. Moreover, several high-performing multimodal methods, including DrFuse, ShaSpec, SMIL, and InfoReg, consistently demonstrate higher gaps across AUPRC, TPR, FPR, and calibration (ECE) compared to the unimodal Transformer. These results suggest that performance gains from multimodal fusion do not automatically translate into improved fairness, and in some cases may even exacerbate subgroup disparities.

\textbf{\textit{Finding 8: Fairness violations are driven more by unequal sensitivity than false positives. }}
Across nearly all models, including the unimodal Transformer, TPR gaps are consistently larger than FPR gaps, indicating that subgroup disparities primarily manifest as unequal sensitivity rather than unequal false positive rates. Although AUPRC and calibration gaps are larger in absolute magnitude, this pattern highlights under-detection of certain demographic groups as a dominant error mode compared to over-detection, with important implications for clinical deployment.

\section{Conclusion}
We introduced CareBench, the first benchmark to understand when multimodal learning helps in clinical prediction using MIMIC-IV and MIMIC-CXR. Our results show that multimodal fusion improves performance when modalities are complete, with gains concentrating in diseases requiring complementary information from both EHR and imaging, but rapidly degrades under modality missingness unless models are explicitly designed. We further find that EHR’s rich temporal structure introduces modality imbalance that architectural complexity cannot overcome. Finally, multimodal fusion does not inherently improve algorithmic fairness, with subgroup disparities primarily arising from unequal sensitivity across demographic groups.

\bibliographystyle{ACM-Reference-Format}
\bibliography{carebench}

\appendix
\section{Data Exclusion Criteria and Data Statistics}
\begin{figure}[H]
    \centering
    \includegraphics[width=0.45\linewidth]{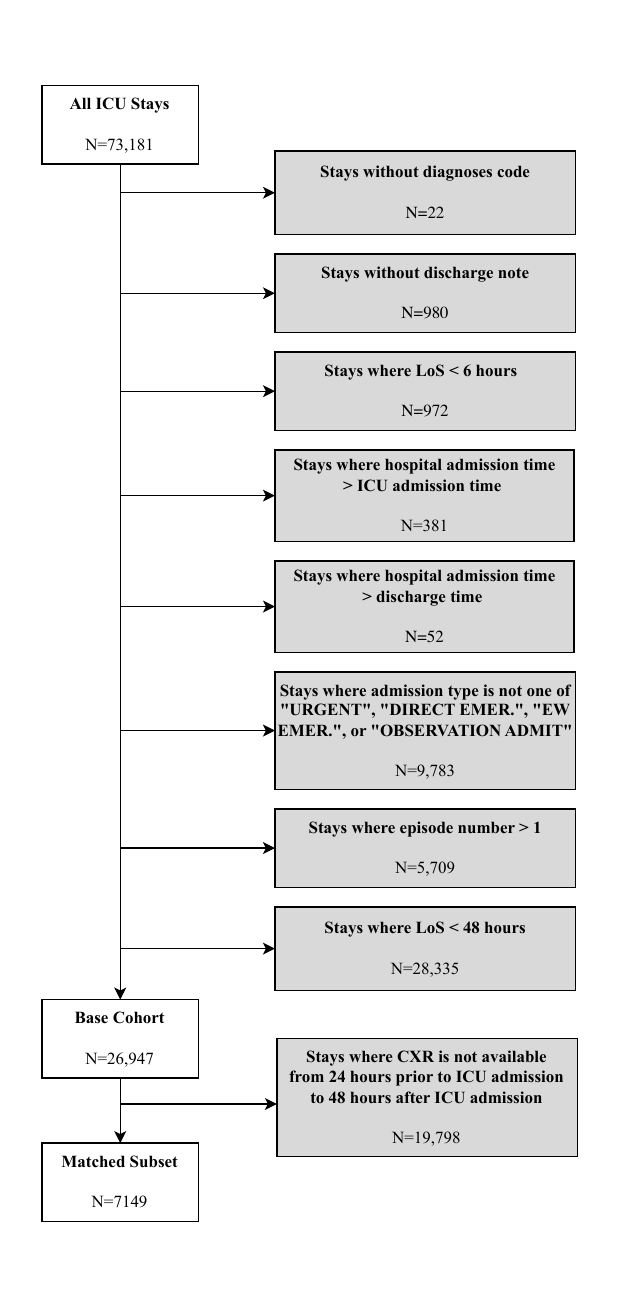}
    \caption{Data exclusion criteria for constructing the base cohort and the matched subset.}
    \label{fig:exclusion_criteria}
\end{figure}

We report our exclusion criteria in constructing the data cohorts in \cref{fig:exclusion_criteria}. The extracted features are summarized in  \cref{tab:features}.

\begin{table}
\centering
\caption{Summary of extracted features from MIMIC-IV derived. Continuous variables are reported as mean$\pm$std [25\%,75\%]; categorical variables as median [25\%,75\%].}\label{tab:search_space_overview}
\label{tab:features}
\resizebox{\linewidth}{!}{
\begin{tabular}{llllc}
\toprule
Category & Feature & MIMIC-IV Source & Summary Stats & Missing \% \\
\midrule
\multicolumn{5}{c}{\textbf{Categorical Features}}\\
\midrule
\multirow{4}{*}{Glasgow Coma Scale} 
  & Eye Opening & gcs.gcs\_eyes & 4 [3,4] & 70.98 \\
  & Verbal Response & gcs.gcs\_verbal & 5 [0,5] & 71.00 \\
  & Motor Response & gcs.gcs\_motor & 6 [6,6] & 71.04 \\ 
  & Total Score & gcs.gcs\_total & 15 [15,15] & 70.93 \\ 
\midrule
\multirow{5}{*}{Cardiac Rhythm} 
  & Absence of Ectopy & mimiciv\_icu.chartevents & binary presence & 30.77 \\
  & Ectopy type: PVCs  & mimiciv\_icu.chartevents & binary presence & 89.76 \\
  & Atrial Fibrillation (AF) & mimiciv\_icu.chartevents & binary presence & 89.91 \\
  & Sinus Rhythm (SR) & mimiciv\_icu.chartevents & binary presence & 43.89 \\
  & Sinus Tachycardia (ST) & mimiciv\_icu.chartevents & binary presence & 85.73 \\
\midrule
\multicolumn{5}{c}{\textbf{Continuous Features}}\\
\midrule
\multirow{7}{*}{Urine Output (KDIGO)} 
  & Urine Output & urine\_output.urineoutput & 106.44$\pm$88.17 [40.00,150.00] & 47.81 \\
  & Urine Output rate 6h  & kdigo\_uo.uo\_rt\_6hr & 0.93$\pm$0.60 [0.48,1.27] & 58.25 \\
  & Urine Output rate 12h  & kdigo\_uo.uo\_rt\_12hr & 0.92$\pm$0.56 [0.50,1.25] & 65.69 \\
  & Urine Output rate 24h  & kdigo\_uo.uo\_rt\_24hr & 0.93$\pm$0.52 [0.53,1.25] & 80.18 \\
  & Observation time 6h  & kdigo\_uo.uo\_tm\_6hr & 6.95$\pm$0.17 [7.00,7.00] & 47.81 \\
  & Observation time 12h  & kdigo\_uo.uo\_tm\_12hr & 11.68$\pm$3.10 [11.00,13.00] & 47.81 \\
  & Observation time 24h  & kdigo\_uo.uo\_tm\_24hr & 16.77$\pm$8.69 [9.00,25.00] & 47.81 \\
\midrule
Oxygen Delivery 
  & O$_2$ Flow & oxygen\_delivery.o2\_flow & 3.08$\pm$1.43 [2.00,4.00] & 89.48 \\
\midrule
Ventilator Setting 
  & Fraction of Inspired Oxygen (FiO$_2$) & ventilator\_setting.fio2 & 46.93$\pm$11.86 [40.00,50.00] & 89.67 \\
\midrule
\multirow{8}{*}{Vital Signs} 
  & Heart Rate & vitalsign.heart\_rate & 84.29$\pm$17.34 [72.00,96.00] & 5.19 \\
  & Respiratory Rate & vitalsign.resp\_rate & 19.12$\pm$4.85 [16.00,22.00] & 5.97 \\
  & Systolic Blood Pressure & vitalsign.sbp & 118.40$\pm$20.21 [103.00,132.00] & 9.04 \\
  & Diastolic Blood Pressure  & vitalsign.dbp & 62.78$\pm$13.61 [53.00,72.00] & 9.06 \\
  & Mean Blood Pressure  & vitalsign.mbp & 77.76$\pm$13.71 [68.00,87.00] & 8.94 \\
  &  Oxygen Saturation (SpO$_2$) & vitalsign.spo2 & 96.82$\pm$2.54 [95.00,99.00] & 7.29 \\
  & Temperature & vitalsign.temperature & 36.87$\pm$0.48 [36.56,37.17] & 71.89 \\
  & Glucose & vitalsign.glucose & 138.36$\pm$44.83 [106.00,163.00] & 79.45 \\
\midrule
Weight 
  & Weight & weight\_durations.weight\_daily & 81.75$\pm$20.28 [66.80,95.00] & 83.78 \\
\bottomrule
\end{tabular}
}
\end{table}

\section{Comparisons with Existing Benchmarks}

\begin{table}[]
    \centering
    \caption{Comparison of CareBench and existing benchmarks.}\label{tab:comparison}
    \resizebox{\linewidth}{!}{
    \begin{tabular}{cccccccc}
    \toprule
      \textbf{Benchmark}  & \textbf{Modalities} & \textbf{Multimodal?} & \textbf{\# of Models} & \textbf{Accuracy?} & \textbf{Robustness?} & \textbf{Fairness?}  \\\midrule
      \citet{purushothamBenchmarkingDeepLearning2018}  & EHR & \redcross & 18 & \greencheck & \redcross  & \redcross \\
      \citet{harutyunyanMultitaskLearningBenchmarking2019} & EHR & \redcross & 7 & \greencheck & \redcross  & \redcross \\
      \citet{barbieriBenchmarkingDeepLearning2020} & EHR & \redcross & 13 & \greencheck & \redcross  & \redcross \\
      MIMIC-Extract~\citep{wangMIMICExtractDataExtraction2020} & EHR & \redcross & 5 & \greencheck & \redcross  & \redcross \\
      ~\cite{sheikhalishahiBenchmarkingMachineLearning2020} & EHR & \redcross & 4 & \greencheck & \redcross  & \redcross \\
      
      FIDDLE~\citep{tangDemocratizingEHRAnalyses2020} & EHR & \redcross & 4 & \greencheck & \redcross  & \redcross \\
      Clairvoyance~\citep{jarrettCLAIRVOYANCEPIPELINETOOLKIT2021} & EHR & \redcross & 7 & \greencheck & \greencheck & \redcross \\
RadFusion~\citep{zhou2021radfusion} & EHR \& CT  & \greencheck & 1 & \greencheck & \redcross & \greencheck \\
EHR-TS-PT~\citep{mcdermott2021comprehensive} & EHR  & \redcross & 1 & \greencheck & \redcross & \redcross \\

      HiRID-ICU~\citep{yecheHiRIDICUBenchmarkComprehensiveMachine2022} & EHR & \redcross & 6 & \greencheck & \greencheck & \redcross \\
      EHRSHOT ~\citep{wornow2023ehrshot} & EHR & \redcross & 2 & \greencheck & \redcross  & \redcross \\

      MC-BEC~\citep{chen2023multimodal} & EHR \& Text \& Waveforms  & \greencheck & 1   & \greencheck & \greencheck & \greencheck \\
      
    INSPECT~\citep{huang2023inspect} & EHR \& CT \& Text & \greencheck & 1 & \greencheck & \redcross & \redcross \\
    MEDFAIR~\citep{zong23medfair} & Imaging & \redcross & 11 & \greencheck & \redcross & \greencheck \\
       YAIB~\citep{vandewater24yet}  & EHR & \redcross & 8 & \greencheck & \redcross  &  \redcross \\
       
       MedMod~\citep{elsharief25medmod} & EHR \& CXR & \greencheck & 11 & \greencheck & \redcross & \redcross\\
 
       CareBench (ours) & EHR \& CXR & \greencheck & 17 & \greencheck & \greencheck & \greencheck \\
         \bottomrule
    \end{tabular}
    }
\end{table}

\section{Implementation Details}
\subsection{Benchmark models}\label{app:models}
We benchmark a broad set of models for multimodal fusion of EHR and chest X-rays, spanning unimodal baselines, simple fusion strategies, and recent state-of-the-art multimodal algorithms that can be adapted to clinical settings.

\paragraph{Uni-modal Baselines} We include uni-modal models as baselines to establish reference performance for each modality. For EHR, we include the classic Long Short-Term Memory network (LSTM) and the Transformer model. They are widely used architectures for capturing temporal dependencies in sequential EHR data. For CXR, we adopt the ResNet-50 model, which is pretrained on ImageNet. These baselines quantify the stand-alone predictive value of each modality.

\paragraph{Complete-Modality Multimodal Fusion Methods}  This group of models assumes that all modalities are present at both training and inference, covering both multimodal fusion architectures and training strategies designed for complete-modality settings, including: 
\begin{itemize}[leftmargin=*, itemsep=3pt, parsep=0pt, topsep=1pt, partopsep=0pt]
    \item \textbf{Late Fusion} is a naive multimodal fusion strategy that concatenates unimodal embeddings followed by a classifier. The EHR and CXR are encoded by a Transformer and a ResNet-50, respectively, with encoders and classifier trained jointly. Despite its simplicity, such late fusion strategies remain widely used in clinical machine learning and serve as strong baselines for comparison against more sophisticated designs.
    \item \textbf{Unified Temporal Discretization Embedding (UTDE)}~\citep{zhang2023improving} is originally designed to handle the irregularity of time series and clinical notes in EHR data. It unifies complementary temporal discretization methods by integrating imputation-based and attention-based interpolation embeddings through a gating mechanism, yielding robust representations of irregular time series. For clinical notes, UTDE casts text embeddings with their note-taking times as irregular sequences and applies a time attention module to capture temporal dynamics. The fusion of time series and notes is then achieved via interleaved self- and cross-attention layers that integrate irregularity across modalities.
    \item \textbf{Dynamic Affine Feature Map Transform (DAFT)}~\citep{daft-polsterl2021combining} is a general-purpose fusion module designed to integrate high-dimensional images with complementary low-dimensional tabular data. DAFT dynamically rescales and shifts convolutional feature maps conditional on tabular inputs, enabling fine-grained interaction between modalities beyond simple concatenation. This mechanism allows clinical variables to modulate intermediate image representations, thereby supporting tighter cross-modal exchange. %
    \item \textbf{Multimodal Transfer Module (MMTM)}~\citep{joze2020mmtm} introduces a lightweight plug-in module for CNN-based intermediate fusion. It performs slow fusion by inserting squeeze-and-excitation units into intermediate levels of unimodal backbones, learning a joint representation that adaptively recalibrates channel-wise features across modalities. %
    \item \textbf{AUG}~\citep{Jiang2025aug} addresses modality imbalance from the perspective of classification ability disproportion. It introduces a boosting-based learning strategy that progressively enhances weaker modalities by jointly optimizing classification errors and residuals across modalities. This iterative procedure helps reduce the performance gap between strong and weak modalities, leading to more balanced multimodal learning.
    \item \textbf{InfoReg}~\citep{huang2025adaptive} focuses on regulating multimodal information acquisition during training. It identifies a critical early training phase in which information-sufficient modalities dominate the learning process and suppress the information acquisition of weaker modalities, and adaptively slows down the learning of information-rich modalities within this window. By temporally regulating information acquisition, InfoReg allows weaker modalities to learn more discriminative representations and promotes balanced multimodal learning.
\end{itemize}

\paragraph{Missing-Modality Multimodal Fusion Methods} We implement a broad set of multimodal fusion methods that could handle missing modalities, covering both models developed specifically for clinical data and models originally proposed in other domains (e.g., video–audio classification) that can be naturally adapted to clinical EHR–CXR fusion. This collection spans diverse design paradigms, including:
\begin{itemize}[leftmargin=*, itemsep=3pt, parsep=0pt, topsep=1pt, partopsep=0pt]
    \item \textbf{Hybrid Early-fusion Attention Learning Network (HEALNet)}~\citep{hemker2024healnet} introduces a multimodal fusion architecture that combines shared and modality-specific parameter spaces within an iterative attention framework. A shared latent bottleneck array is propagated and updated across layers to capture cross-modal interactions and shared information. In parallel, modality-specific attention weights are learned and reused across layers, enabling the model to preserve structural information unique to each modality while maintaining efficient fusion through shared parameters.
    \item \textbf{Flexible Mixture-of-Experts (Flex-MoE)}~\citep{yun2024flex} is designed to support arbitrary combinations of input modalities without retraining. It constructs a shared latent space where modality-specific encoders map their features, and employs a mixture-of-experts (MoE) fusion layer that dynamically activates experts depending on the available modalities. This enables the model to flexibly integrate any subset of modalities during inference to enhance robustness to missing data and scalability to new modality combinations.
    \item \textbf{DrFuse}~\citep{yao2024drfuse} is a clinical multimodal fusion method proposed for EHR and chest X-ray images. It tackles two key challenges, namely the missing modalities and modal inconsistency. It disentangles shared information (common across EHR and CXR) from modality-specific features and aligns the shared representations via distribution matching. This allows robust inference even when one modality is absent. To further handle patient- and disease-specific variability, DrFuse introduces a disease-aware attention fusion module that adaptively weights each modality. Following its original settings, we adopt Transformer and ResNet-50 as the encoders for EHR and CXR, respectively.
    \item \textbf{Unified Multi-modal Set Embedding (UMSE)}~\citep{lee2023learning} addresses the irregular sampling and missing modalities in multimodal EHR learning. It encodes values, time, and feature types across all modalities within a shared embedding framework. By sharing the time and feature embeddings, UMSE preserves temporal relationships between heterogeneous modalities without relying on carry-forward or imputation. To tackle the missing modalities, a Skip Bottleneck (SB) is introduced to enable the Multimodal Bottleneck Transformer to process data with missing modality.
    \item \textbf{Shared-Specific Feature Modelling (ShaSpec)}~\citep{wang2023multi} is a multimodal learning framework that decomposes each modality into shared features that are modality-robust and specific features that capture modality-unique information. These components are combined through a residual fusion mechanism. To enforce disentanglement, ShaSpec applies distribution alignment on shared features and a domain classification objective on modality-specific features.
    \item \textbf{M3Care}~\citep{zhang2022m3care} addresses the challenge of missing modalities in multimodal healthcare data. Instead of generating raw missing data, it imputes task-relevant latent representations by leveraging auxiliary information from clinically similar patients. Specifically, M3Care employs task-guided modality-adaptive kernels to construct patient similarity graphs, aggregates information from these neighbors, and adaptively fuses it with available modalities.
    \item \textbf{MedFuse}~\citep{mlhc2022hayatmedfuse} is a multimodal fusion method tailored for EHR and chest X-ray images, particularly focusing on missing modalities. After obtaining representations of each modality, MedFuse treats the modality-specific representations (EHR and CXR) as a sequence and aggregates them with an LSTM-based fusion module. This recurrent design enables the model to naturally handle missing modalities by processing variable-length input sequences. Compared to conventional early or joint fusion, MedFuse improves performance on EHR–CXR prediction tasks while maintaining robustness under partial modality availability.
    \item \textbf{SMIL}~\citep{ma2021smil} addresses multimodal learning when a large fraction of training and testing samples lack one or more modalities. It introduces a Bayesian meta-learning framework that contains three components: (i) a feature reconstruction network that approximates missing modality features conditioned on observed ones, (ii) a feature regularization network that perturbs latent embeddings to mitigate bias from incomplete data, and (iii) a main prediction network. This unified design aims to handle different missing-modality patterns during both training and inference, and to train efficiently when most samples are incomplete.
\end{itemize}

\end{document}